%% file: acl_latex.tex
\pdfoutput=1
\documentclass[11pt]{article}

\usepackage[final]{acl} 

\usepackage{times}
\usepackage{latexsym}
\usepackage[T1]{fontenc}
\usepackage[utf8]{inputenc}
\usepackage{microtype}
\usepackage{inconsolata}
\usepackage{graphicx}
\usepackage{amsmath}
\usepackage{algorithm}
\usepackage{algpseudocode}
\usepackage{float}
\usepackage{stfloats}
\usepackage{geometry}
\usepackage{adjustbox}
\usepackage{enumitem}
\usepackage{booktabs}
\usepackage{tcolorbox}
\usepackage{setspace}
\usepackage{multirow}
\usepackage{makecell}
\usepackage{rotating}
\usepackage{array}
\usepackage{xcolor}
\usepackage{colortbl}
\usepackage{arydshln}
\usepackage{subcaption}

\usepackage{kotex}

\title{Pre-Storage Reasoning for Episodic Memory: \\Shifting Inference Burden to Memory for Personalized Dialogue}

\author{
    \textbf{Sangyeop Kim\textsuperscript{1,2}$^*$},
    \textbf{Yohan Lee\textsuperscript{3}\thanks{These authors contributed equally.}},
    \textbf{Sanghwa Kim\textsuperscript{4}},
    \textbf{Hyunjong Kim\textsuperscript{1}},
    \textbf{Sungzoon Cho\textsuperscript{1}\thanks{Corresponding author.}}
    \vspace{0.4em}
    \\
    \normalsize{
        \textsuperscript{1}Seoul National University,
        \textsuperscript{2}Coxwave,
        \textsuperscript{3}Independent Researcher,
        \textsuperscript{4}KAIST
    }
    \vspace{0.2em}
    \\
    \normalsize{
        \texttt{
        sy917kim@bdai.snu.ac.kr, yhlee.nlp@gmail.com, zoon@snu.ac.kr}
    }
}

\begin{document}
\maketitle
\begin{abstract}
Effective long-term memory in conversational AI requires synthesizing information across multiple sessions. However, current systems place excessive reasoning burden on response generation, making performance significantly dependent on model sizes. We introduce PREMem (Pre-storage Reasoning for Episodic Memory), a novel approach that shifts complex reasoning processes from inference to memory construction. PREMem extracts fine-grained memory fragments categorized into factual, experiential, and subjective information; it then establishes explicit relationships between memory items across sessions, capturing evolution patterns like extensions, transformations, and implications. By performing this reasoning during pre-storage rather than when generating a response, PREMem creates enriched representations while reducing computational demands during interactions. Experiments show significant performance improvements across all model sizes, with smaller models achieving results comparable to much larger baselines while maintaining effectiveness even with constrained token budgets. Code and dataset are available at \url{https://github.com/sangyeop-kim/PREMem}.
\end{abstract}

\section{Introduction}
Human cognition seamlessly synthesizes past experiences into coherent episodic memories that support personalized interactions \cite{piaget1952origins, carey1985conceptual, 10.5555/2222503}. When engaging with familiar people, individuals effortlessly perform relevant interactions, track evolving preferences, and maintain consistent mental models without explicitly reviewing conversation histories. This natural memory process enables meaningful relationships through contextualized understanding.

In conversational AI, well-designed memory structures are essential for maintaining personalized interactions across multiple sessions \cite{martins-etal-2022-infinity-former, bae-etal-2022-keep-me-update, gutierrez2024hipporag}. Effective memory mechanisms allow AI assistants to track user preferences, recall shared experiences, and sustain consistent understanding over time—capabilities that form the foundation of truly personalized dialogue systems \cite{wu2025humanmemoryaimemory, fountas2025humaninspired}.

Current memory approaches in conversational AI systems rely on three core mechanisms \cite{wang2024towards, du2025rethinkingmemoryaitaxonomy}: indexing and storing, retrieval, and memory-based generation. Recent advances have explored various structural granularities---from turn-level and session-level segmentation to compressed summaries \cite{pan2025secom} and knowledge graphs \cite{edge2025localglobalgraphrag, zhu2025knowledgegraphguidedretrievalaugmented}. These approaches primarily investigate how different memory structures affect retrieval efficiency and accuracy, yet struggle with cross-session challenges that require understanding continuity, causality, and state changes.

Recent works \cite{xu2025a-mem, gutierrez2025hipporag2} have attempted to address multi-session reasoning through metadata annotations and concept-linking knowledge graphs. However, these methods typically define cross-session relationships as simple clusters without modeling the nature of relationships or temporal evolution.

Beyond these limitations of retrieval-focused approaches, a more critical challenge emerges even when retrieval succeeds. Even with optimal retrieval systems that can provide relevant context, models frequently struggle with complex reasoning tasks that require synthesis and inference—particularly temporal relationships and cross-session information integration \cite{10.1145/3477495.3531961, yuan-etal-2025-personalized}. Unlike simple information retrieval, these tasks demand sophisticated cognitive processes including pattern recognition, causal reasoning, and contextual synthesis. This computational burden during response generation creates significant inefficiency and amplifies performance disparities between large and small models.

To address these challenges, we present \textbf{PREMem} (\textbf{P}re-storage \textbf{R}easoning for \textbf{E}pisodic \textbf{Mem}ory), a cognitive science-grounded approach that shifts complex reasoning processes from response generation to memory construction. Our approach draws inspiration from human cognitive processes: rather than exhaustively reviewing conversation histories during interactions, humans rely on pre-consolidated memories that have undergone sophisticated synthesis during offline periods \cite{squire1987memory, schacter2007constructive}. Based on schema theory \cite{rumelhart1976accretion, bartlett1995remembering}, human memory actively transforms information during storage through assimilation and accommodation processes, enabling efficient retrieval and coherent understanding across temporal contexts.

\begin{figure*}[!thbp]
    \centering
    \includegraphics[width=\linewidth]{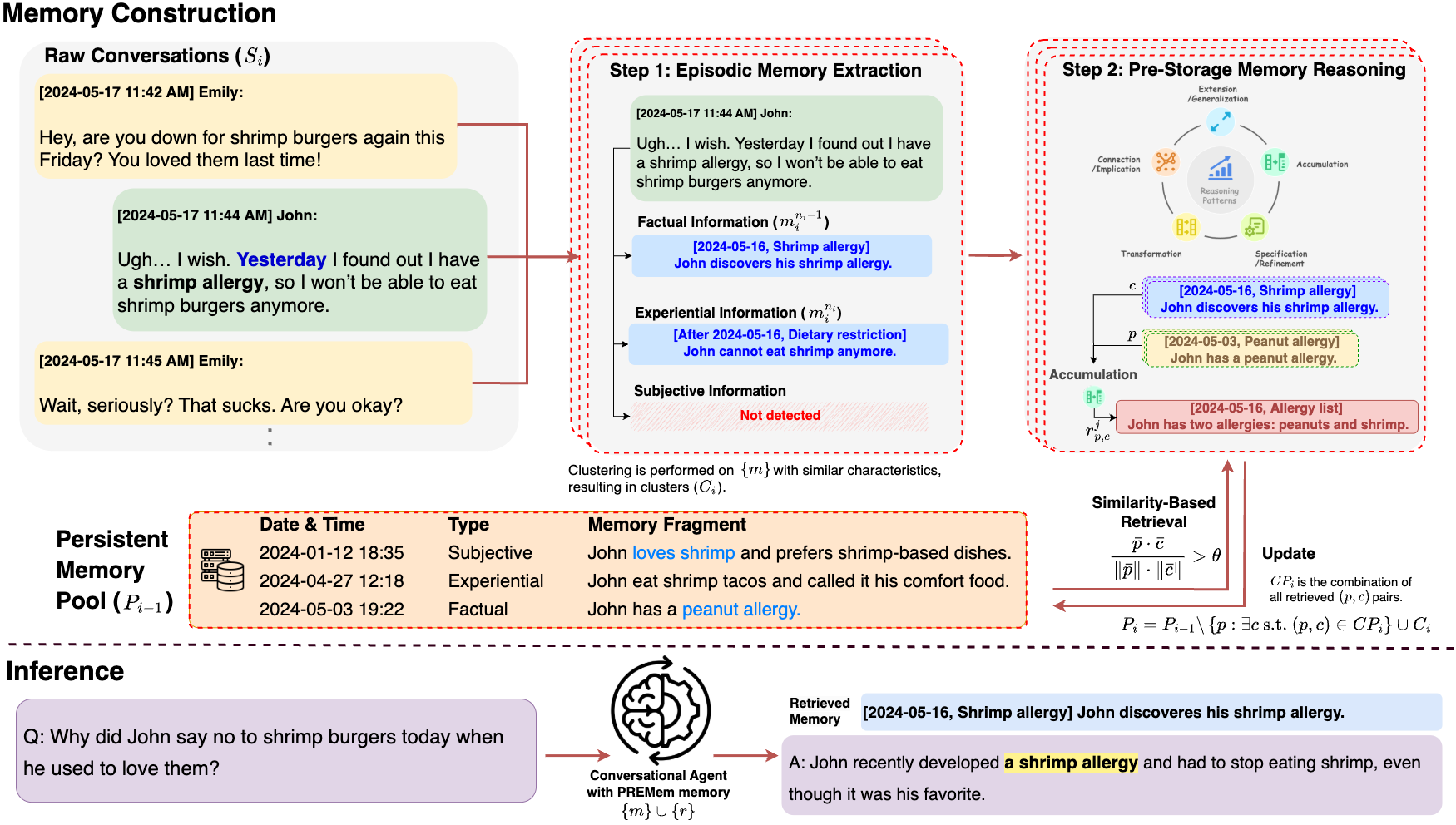}
    \caption{\textbf{PREMem} architecture divided into \textit{Memory Construction} phase (comprising Step 1: Episodic Memory Extraction and Step 2: Pre-Storage Memory Reasoning) and \textit{Inference} phase.}
    \label{fig:main}
\end{figure*}

PREMem implements this cognitive principle by extracting memory fragments into three theoretically-grounded categories—factual, experiential, and subjective information—and establishing explicit cross-session relationships through five evolution patterns derived from schema modification mechanisms, as shown in Figure \ref{fig:main}. By performing complex reasoning during pre-storage rather than at response time, our approach creates enriched memory representations while reducing computational demands during interactions—offering both performance gains and practical deployment advantages.

Experimental results on LongMemEval \cite{wu2025longmemeval} and LoCoMo \cite{maharana-etal-2024-evaluating-locomo} benchmarks demonstrate significant improvements across all model sizes. PREMem shows particularly strong results on cross-session reasoning tasks, with even small language models (≤4B) achieving competitive performance compared to much larger baseline models. Additional experiments confirm its practical applicability in resource-constrained environments through efficient token utilization.

Our contributions include: (1) A cognitive science-grounded memory framework based on established schema theory that extracts structured episodic fragments and models information evolution through five theoretically-validated patterns; (2) A pre-storage reasoning method that shifts complex cross-session synthesis from response time to memory construction, mirroring human cognitive consolidation processes; (3) Comprehensive experimental validation across two benchmarks, multiple model families and question types, demonstrating robust generalization; (4) Practical advantages for resource-constrained applications through reduced inference-time computational requirements.

\section{Related Works}

\subsection{Memory in Conversational AI Systems}
Long-term memory in conversational AI systems requires integrating and updating experiences across multi-turn dialogues~\cite{wang2024towards, du2025rethinkingmemoryaitaxonomy}. Existing approaches employ unstructured formats such as summarization~\cite{Zhong_Guo_Gao_Ye_Wang_2024, WANG2025130193} or compression~\cite{pan2025secom, chen-etal-2025-compress}, but struggle with temporal modeling and content overlap, leading to information loss and fragmented representations.

Knowledge graph-based methods~\cite{edge2025localglobalgraphrag, guo2025lightragsimplefastretrievalaugmented, zhu2025knowledgegraphguidedretrievalaugmented} enhance semantic connectivity through structured representations, but their partial graph construction prevents establishing relationships between temporally distant nodes across conversation sessions.

Recent efforts such as \citet{li-etal-2025-hello} and \citet{ong-etal-2025-towards} introduce modular memory architectures and timeline-based linking to better reflect dialogue dynamics. However, these approaches still perform memory relationship reasoning during response generation, making them heavily dependent on the capabilities of the underlying model.

Recent systems~\cite{pmlr-v235-lee24c-readagent, xu2025a-mem, yuan-etal-2025-personalized} support dynamic memory evolution and attempt to establish connections between memories. However, they rely on implicit, unstructured associations rather than explicit schemas for modeling information evolution across sessions. This approach can lead to arbitrary links and inconsistent interpretations that are difficult to analyze.

We address these limitations with PREMem, a novel structured memory approach. Our method provides clear temporal relationships, well-defined semantic connections between related information, and systematically organized memory representations that enhance consistency, interpretability, and reasoning efficiency.

\subsection{Cognitive Perspectives on Memory}

Memory in AI-based conversational systems shares structural and functional characteristics with human memory, prompting researchers to incorporate cognitive science principles into memory system design \cite{wang2024userbehaviorsimulationlarge, shan2025cognitivememorylargelanguage}. This enables systems to maintain consistent user representations across multiple conversations.

Inspired by these cognitive principles, researchers have developed various methods for transforming conversation data into structured episodic memories \cite{10.1145/3613905.3650839, fountas2025humaninspired, ge2025tremuneurosymbolictemporalreasoning}. \citet{10.1145/3613905.3650839} models human memory consolidation by weighting information based on contextual relevance and recall frequency, while \citet{fountas2025humaninspired} applies event cognition principles to segment conversations using prediction errors and graph-theoretical clustering.

However, these approaches face limitations in cross-session reasoning, as they focus more on storage organization than on modeling information evolution across conversations \cite{qiu-etal-2024-large, chu-etal-2024-timebench}. Although systems like \citet{xu2025a-mem} and \citet{gutierrez2025hipporag2} attempt to address this through linked structures, they still struggle with tracking changing preferences and resolving contradictions \cite{huet2025episodic, wu2025longmemeval}.

To overcome these limitations, we examine how humans reason about and synthesize memories. Cognitive science offers guidance through schema theory—detailed in Appendix \ref{apdx:schema}. This theory views memory as a structured interpretive system \cite{piaget1952origins, rumelhart1976accretion, bartlett1995remembering, rumelhart2017schemata}. In this framework, new information actively integrates with existing knowledge through generalization and exception handling \cite{fauconnier2008way, chi2009three}.

Based on these insights, our study not only structures conversations into temporal episodic units but also models the semantic relationships between them. This approach captures continuity, causality, and change patterns across conversations, enabling more consistent and personalized responses even as user preferences evolve over time.

\section{Methodology}

We present \textbf{PREMem}, a novel approach that shifts complex memory synthesis and analysis from response generation to the memory construction phase. By performing pre-storage reasoning across conversations, our approach reduces the computational burden during dialogue while creating more cognitive-inspired memory representations. 
Figure \ref{fig:main} illustrates the overall architecture of our approach, which consists of a \textit{Memory Construction} phase (with two steps detailed in the following sections) and an \textit{Inference} phase. 
This method improves personalized conversation performance across all model sizes, with smaller models (≤4B) achieving results comparable to baselines using much larger models. 
All prompts and pseudo code can be found in Appendix \ref{apdx:llm_prompt} and \ref{apdx:pseudo_code}, respectively.

\subsection{Step 1: Episodic Memory Extraction}
We extract episodic memory from conversation history, classifying it into three categories that reflect human memory components \cite{squire1987memory, schacter1994memory}:

\begin{enumerate}[label={\tiny$\bullet$}, leftmargin=15pt, itemsep=5pt, topsep=5pt, parsep=3pt, partopsep=5pt]
    \item \textbf{Factual Information}: Objective facts about personal states, attributes, possessions, and relationships (``what I am/have/know'')
    \item \textbf{Experiential Information}: Events, actions, and interactions experienced over time (``what I did/experienced'')
    \item \textbf{Subjective Information}: Internal states including preferences, opinions, beliefs, goals, and plans (``what I like/think/want'')
\end{enumerate}

Beyond comprehensive categorization, effective memory structure needs to solve the challenge of \textit{temporal reasoning}—determining accurate time relationships. Previous research \cite{xu2025a-mem} shows language models struggle with relative time expressions such as ``yesterday'' and ``last week''. We address this through a structured temporal representation with four patterns: (1) ongoing facts use message dates directly; (2) specific past events convert relative expressions to absolute dates; (3) unclear past events use ``Before [message-date]''; and (4) future plans use ``After [message-date].''

We formalize memory extraction through $LLM_{extract}$ which operates on conversation sessions $S_1, S_2, \cdots, S_N$:

\vspace{-1em}
$$LLM_{extract}(S_i) \rightarrow \{m_i^1, m_i^2, ..., m_i^{n_i}\},$$

where $n_i$ is the number of memory fragments in session $S_i$. Each memory fragment $m_i^j$ includes source identification, key phrase, memory content, and temporal context: 

\vspace{-1em}
$$m_i^j = (\text{id}_i^j, \text{key}_i^j, \text{content}_i^j, \text{time}_i^j).$$

\subsection{Step 2: Pre-Storage Memory Reasoning}
From memory fragments, we analyze relationships between information across conversation sessions using cognitive schema theory \cite{rumelhart1976accretion, anderson2013architecture, meylani2024innovations}. This approach shifts complex cognitive tasks—including pattern recognition, information synthesis, and contextual reasoning—to the storage phase, reducing computational demands during dialogue while creating enriched memory representations with inferred relationships and implications.

\subsubsection{Clustering and Temporal Linking}

We organize memory fragments into semantic clusters using embeddings generated from combined key phrases and memory content. For each session $S_i$, we embed the memory fragments $\{ m_i^j \}_{j=1}^{n_i}$ into vectors $\{e_i^j\}_{j=1}^{n_i}$ using an embedding model $f_{emb}$, that is, $e_i^j = f_{emb}(m_i^j)$. Using silhouette scores to determine optimal groupings, we form clusters $C_i = \{c_i^1, c_i^2, ..., c_i^{k_i}\}$ with each cluster containing embedding of semantically related memory items. This clustering step serves two critical purposes: it reduces redundancy in memory representations to minimize noise during reasoning \cite{pan2025secom}, and it prevents combinatorial explosion by limiting the number of cross-session comparisons required during relationship analysis.

For a cluster $c$, the centroid is calculated as $\overline{c} = \frac{1}{|c|} \sum_{e \in c} e$ and the collection of memory fragments corresponding to the cluster $c$ is denoted as $M_{c}$. 

We maintain a persistent memory pool $P_i$ of clusters that have not yet found a semantic match with a cluster that comes after themselves up to the $i$-th session, initialized as $P_0 = \{\}$. For each new session $S_i$, we measure the similarity between existing persistent cluster $p \in P_{i-1}$ and new cluster $c \in C_i$ using the cosine similarity of centroids:

\vspace{-0.5em}
$$sim(p, c) = \frac{\overline{p} \cdot \overline{c}}{||\overline{p}|| \cdot ||\overline{c}||}.$$

We define a pair $(p,c)$ as \textit{connected} if $sim({p}, {c}) > \theta$.
We define a set $CP_i$ that contains connected pairs $(p, c)$, that is,

\vspace{-0.2em}
$$CP_i := \{ (p,c) : sim(p,c) > \theta \}$$
$$\text{where } p \in P_{i-1}, c \in C_i$$

The set $CP_i$ consists of semantically related cluster pairs across sessions.

\subsubsection{Cross-Session Reasoning Patterns}

For each identified connection, we perform cross-session reasoning based on five information evolution patterns derived from schema modification mechanisms \cite{rumelhart1976accretion, anderson2013architecture}. These patterns synthesize findings from extensive cognitive science literature \cite{bransford1972contextual, chi1981expertise, murphy2004big, chi2009three} to capture fundamental ways humans integrate new information with existing knowledge structures. The detailed theoretical foundations are provided in Appendix \ref{apdx:schema}.

\begin{enumerate}[label={\tiny$\bullet$}, leftmargin=15pt, itemsep=3pt, topsep=5pt, parsep=3pt, partopsep=5pt]
    \item \textbf{Extension/Generalization}: Expanding scope of existing information (e.g., inferring broader food preferences from restaurant choices)
    
    \item \textbf{Accumulation}: Reinforcing knowledge through repeated similar information (e.g., recognizing consistent exercise habits)
    
    \item \textbf{Specification/Refinement}: Developing more detailed understanding (e.g., clarifying music preferences from general to specific)
    
    \item \textbf{Transformation}: Capturing changes in states or preferences (e.g., identifying shifts in product satisfaction)
    
    \item \textbf{Connection/Implication}: Discovering relationships between separate information (e.g., linking language study with travel plans)
\end{enumerate}

The model $LLM_{reason}$ generates reasoning memory fragments by analyzing memory fragments in $M_p$ and $M_c$ for $(p,c) \in CP_i$ individually, extracting insights about the evolution patterns:

\vspace{-1em}
$$LLM_{reason}(M_p, M_c) \rightarrow \{r_{p,c}^j \}_{j=1}^{d_{p,c}},$$
where $r_{p,c}^j$ is the reasoning memory fragment that follows the same structure as memory fragments.
We define a reasoning memory pool $R_i$ as the union of reasoning memory fragments $\{r_{p,c}^j \}_{j=1}^{d_{p,c}}$ over all connected pairs $(p,c) \in CP_i$ 
and denote embedding of $R_i$ using embedding model $f_{emb}$ as $E_i'$.

After reasoning on the pair $(p,c)\in CP_i$, we remove $p$ from the persistent memory pool since it finds a semantic match with later-coming cluster $c$. On the other hand, we put all latest clusters $c \in C_i$ into the pool, then we get the updated persistent memory pool $P_i$, which is formally defined as:

\vspace{-1em}
$$P_i = P_{i-1} \setminus \{p : \exists c  \;\; s.t. \;\; (p,c) \in CP_i\} \cup C_i.$$

This process serves two important purposes: first, it prevents computational explosion as sessions increase by eliminating already-processed information; second, it enables efficient long-term topic tracking across temporally distant conversations.

After this whole process is performed on the last conversation session $S_N$, we prepare memory storage $\mathcal{M}$ and reasoning memory storage $\mathcal{R}$ used in inference as $\mathcal{M} := \cup_{i=1}^{N} \{m_i^j\}_{j=1}^{n_i}$ and $\mathcal{R} := \cup_{i=1}^{N} R_i$; and denote their embeddings using $f_{emb}$ as $E$ and $E'$, respectively.

\subsection{Inference Phase}
For a user query ($q$), we retrieve the most relevant items from our total memory store $\mathcal{M} \cup \mathcal{R}$ and select the top-k items based on the similarity between embedded vectors $e \in (E \cup E')$ and $f_{emb}(q)$. These retrieved memory items denoted by $m_*^1, \cdots, m_*^k$ are arranged chronologically and composed to form the \textit{context}, with each item including its complete information (key, content, time). We then generate a response using this organized context:

\vspace{-1.3em}
$$LLM_{response}(\textit{context}, q) \rightarrow response.$$


\section{Experiments}
\subsection{Experimental Setup}
\input{table/data_category_stats}

\paragraph{Datasets}
We utilize two long-term memory QA datasets: LoCoMo \cite{maharana-etal-2024-evaluating-locomo} and LongMemEval \cite{wu2025longmemeval}. LoCoMo contains 1,986 QA instances from conversation history sets, averaging 27.2 dialogues per set with 21.6 turns per dialogue. LongMemEval has 500 QA pairs. We adopt the $\text{LongMemEval}_{\text{S}}$ subset, which reflects more realistic constraints. $\text{LongMemEval}_{\text{S}}$ averages 115K tokens per question.

We unify the question types across both datasets into five categories: \textit{single-hop}, \textit{multi-hop}, \textit{temporal reasoning}, \textit{adversarial}, and \textit{knowledge update} (only in LongMemEval). Detailed dataset statistics for each category are provided in Table~\ref{tab:dataset-stats}, and comprehensive information about the datasets, including unification criteria, is described in Appendix~\ref{apdx:dataset}.

To ensure a fair comparison across models and settings, we standardize the answer generation prompt for all experiments. The specific prompts used for each dataset are shown in Appendix~\ref{apdx:llm_prompt}.

\paragraph{Evaluation Metrics}
We evaluate using BLEU-1, ROUGE-1, ROUGE-L, METEOR, BERTScore, and LLM-as-a-judge score. BLEU-1 measures n-gram precision while ROUGE metrics assess lexical overlap through n-grams. METEOR and BERTScore capture semantic similarity beyond exact matches. LLM-as-a-judge score assesses overall response quality including coherence and informativeness, critical for LongMemEval and LoCoMo tasks that require recalling information from past interactions. For adversarial QA categories, we report accuracy based on the proportion of safe responses that identify unanswerable queries.

\input{table/main_table_rank}

\paragraph{Baselines}
We compare our approach against baselines with varying memory granularity and state-of-the-art models. For granularity, we implement turn-level and session-level memory structures. For advanced approaches, we evaluate SeCom~\cite{pan2025secom}, which partitions dialogue into topic-based segments with compression-based denoising; HippoRAG-2~\cite{gutierrez2025hipporag2}, which encodes memory as an open knowledge graph with concept-context structures; and A-Mem~\cite{xu2025a-mem}, which organizes interconnected, evolving notes with semantic metadata.

\paragraph{Implementation Details}
In PREMem, we use identical LLMs for extraction and reasoning, using the largest variant per family: Qwen2.5-72B, Gemma3-27B, or gpt-4.1-base (``base'' distinguishes from smaller variants). For $LLM_{response}$, we evaluate across three LLM families—gpt-4.1 \cite{openai2025gpt41} (nano, mini, base), Qwen2.5 \cite{qwen2.5} (3B, 14B, 72B), and Gemma3 \cite{gemmateam2025gemma3technicalreport} (4B, 12B, 27B)—to assess generalizability across different model capacities. During response generation, all models operate with a temperature of $0.7$. LLM-as-a-judge score uses a deterministic decoding (temperature $0.0$). We use Stella\_en\_400M\_v5~\cite{zhang2025jasperstelladistillationsota} as the embedding model to encode memory items and queries during retrieval. Additional implementation details are provided in Appendix \ref{apdx:implementation}.

\subsection{Main Results}

Table \ref{tab:main_results} shows comprehensive results across LongMemEval and LoCoMo benchmarks using LLM-as-a-judge scores and ROUGE-1. PREMem achieves superior performance across most categories and model sizes, especially in complex reasoning tasks. For overall performance, PREMem consistently outperforms all baselines by substantial margins across both benchmarks.

Results highlight two key findings. First, PREMem demonstrates exceptionally strong performance on challenging cross-session reasoning tasks—multi-hop questions, temporal reasoning, and knowledge update categories. Second, while some baselines excel in specific subcategories (e.g., A-Mem on single-hop questions), PREMem delivers more consistent performance enhancement across all question types, maintaining robust results regardless of question complexity.

\subsection{Small Language Models}

\vspace{-0.4em}
\input{table/small_model_1_3}

\input{table/Ablation_table}
Table~\ref{tab:small} shows LLM-as-a-judge scores comparing PREMem with small models against baseline methods using larger models. The results demonstrate that PREMem enables competitive performance even under limited model capacity.

In Gemma and gpt families, PREMem with smaller models outperforms baselines using larger counterparts across both benchmarks. For the Qwen family, all memory methods using Qwen2.5-3B achieve scores below 50 on both benchmarks, except for PREMem which reaches 50.8 on LongMemEval. With Qwen2.5-14B (Table~\ref{tab:main_results}), PREMem performance surpasses all baseline methods that use the much larger 72B model on both benchmarks. By offloading complex reasoning to the storage phase, PREMem enhances lightweight models with rich memory representations, reducing reliance on large-scale inference models.

\subsection{Ablation Study}

Table \ref{tab:ablation} shows ablation studies of PREMem. Step 1 (memory extraction) is vital, as its removal drops scores by 32.7-69.0\%; similarly, Step 2 (pre-storage reasoning) proves valuable through cross-session pattern analysis. Our episodic memory categorization and temporal reasoning also contribute meaningfully, with their removal decreasing scores by up to 8.9\% and 16.4\% respectively.

These results confirm two key insights. First, our structured approach for memory extraction effectively organizes user information into meaningful categories. Second, performing cross-session reasoning before retrieval time significantly enhances performance across all model sizes. By shifting complex cognitive processes to the memory construction phase, models can focus on response generation during inference, leading to more effective handling of temporal relationships and multi-session information synthesis.

\section{Practical Applications}

Memory systems are foundational for personalized conversational agents, with resource efficiency critical for real-world deployment. To demonstrate the value of PREMem under resource constraints, we evaluate three key dimensions: (1) storage efficiency through alternative retrieval methods (Section \ref{bm25_alternative}), (2) computational cost reduction using smaller reasoning models (Section \ref{smaller_reasoning_model}), and (3) token budget for context efficiency (Section \ref{token_budget_context}).

\subsection{BM25 as Embedding Alternative}
\label{bm25_alternative}

\begin{figure}[!htbp]
  \centering
  \includegraphics[width=\columnwidth]{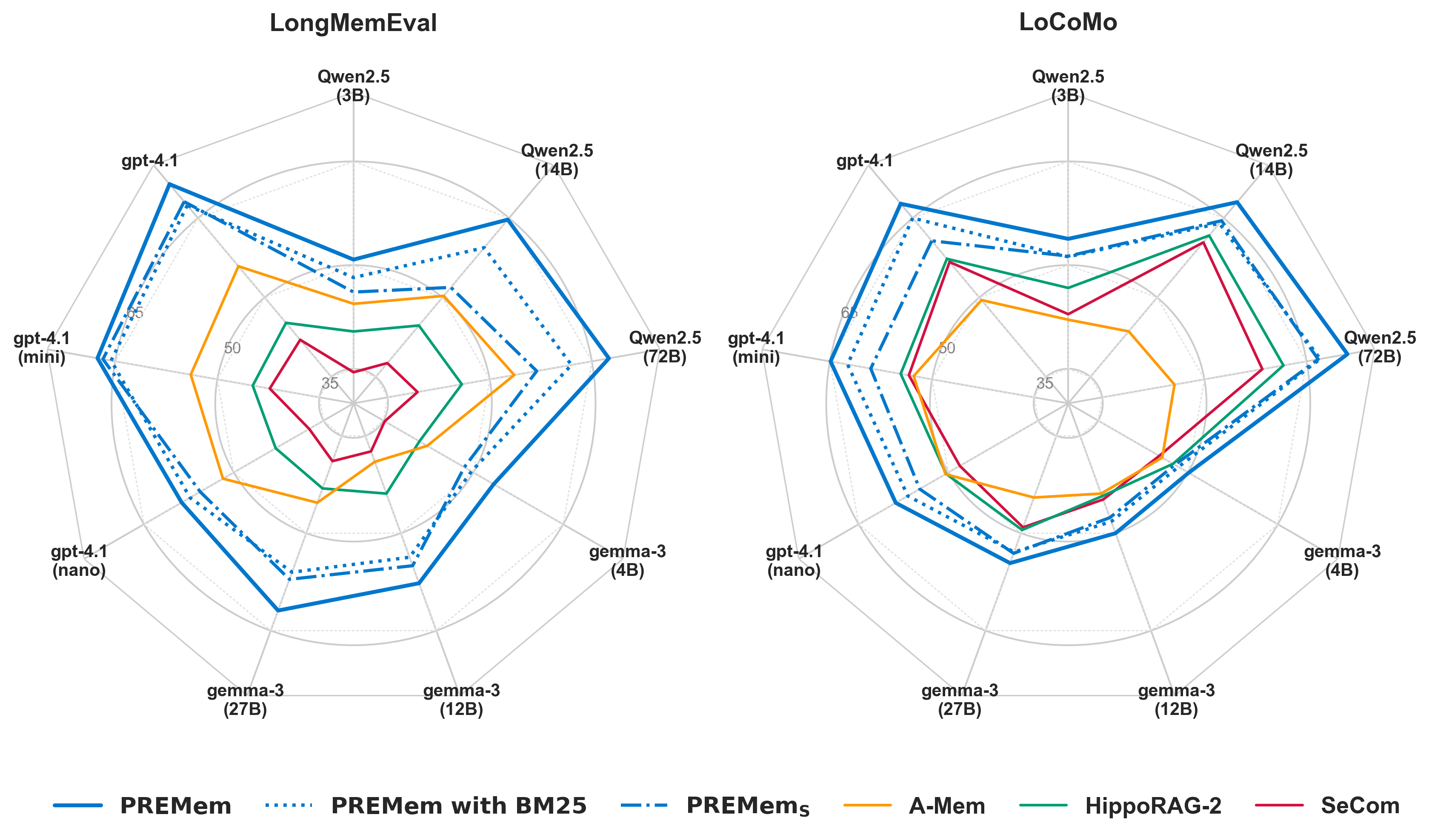}
  \caption{Performance comparison (LLM-as-a-judge score) across retrieval mechanisms (left: BM25 vs. embedding) and memory reasoning models (right: low-spec vs. high-spec).}
  \label{fig:bm25-low-spec}
\end{figure}

Vector embeddings for semantic search demand substantial storage for personalized assistants that must maintain separate indexes for each user. While keyword-based retrieval methods like BM25 typically underperform semantic search methods \cite{thakur2021beir}, experimental results shown in Figure \ref{fig:bm25-low-spec} (left) demonstrate BM25 remain surprisingly competitive with PREMem. This provides an efficient option for resource-constrained deployments with minimal performance tradeoffs.

\subsection{Low-Spec Reasoning Models}
\label{smaller_reasoning_model}
Memory construction typically requires powerful LLMs. To explore more efficient alternatives, we investigate whether smaller models can effectively perform our pre-storage reasoning. We introduce $\text{PREMem}_{\text{S}}$, which uses smaller variants from each model family (Qwen2.5-14B, Gemma3-12B, or gpt-4.1-nano) for memory construction.

Figure~\ref{fig:bm25-low-spec} (right) shows that reasoning-focused prompts in $LLM_{extract}$ and $LLM_{reason}$ help smaller models create high-quality memory representations. This approach effectively provides an alternative for real-world applications by reducing computational costs during memory construction.

\subsection{Token Budget Efficiency}
\label{token_budget_context}

Allocating thousands of tokens from limited context windows solely for memory retrieval represents a substantial opportunity cost for multi-purpose assistants. We evaluate PREMem performance across varying token budgets.

\input{table/token_budget}

While other methods degrade significantly with reduced context lengths, PREMem maintains robust performance even with minimal token, as shown in Table \ref{tab:token-budget}. This stability stems from memory fragments that capture pre-reasoned cognitive relationships rather than simply storing raw conversation turns or graph connections. The efficiency allows developers to allocate smaller portions of context windows to memory while preserving personalization quality in real-world applications.

\section{Conclusion}
We present PREMem, a novel episodic memory system that shifts complex reasoning processes from response generation to the memory construction phase. This method transforms conversations into structured memories with categorized information types and cross-session reasoning patterns. Our approach significantly improves performance on LongMemEval and LoCoMo benchmarks, with particularly strong results for temporal reasoning and multi-session tasks. Notably, even modest-sized models using PREMem achieve competitive results compared to larger state-of-the-art systems. Additionally, our focus on token budget, retrieval efficiency, and streamlined memory construction makes PREMem effective for real-world conversational AI systems that require long-term personalization under resource constraints.

\section*{Limitations}
Our work has several limitations that present opportunities for future research:

\paragraph{Reduced efficiency in single-hop reasoning} Our pre-reasoning structure shows lower performance for single-hop questions compared to direct retrieval methods. This could be due to additional processing that may not benefit straightforward queries. To address this, future work could consider utilizing original messages directly for single-hop reasoning tasks.

\paragraph{Lack of original conversation context} Our implementation focuses on extracted and synthesized memory items rather than original conversation messages to reduce storage requirements. This approach sacrifices access to certain linguistic nuances, including users' conversational styles and terminology preferences. A potential solution might involve query-dependent hybrid retrieval that combines structured memories with original conversation segments based on the nature of the user's question.

\paragraph{Absence of memory decay mechanisms} Our approach does not incorporate forgetting mechanisms found in human memory. While our similarity threshold helps filter retrieved items, managing truly long-term conversations would require additional constraints. For extended conversation histories, implementing explicit memory size limitations or importance-based decay functions would help control the persistent memory pool.

\paragraph{Limited theoretical contribution} Our approach demonstrates practical improvements by applying cognitive science concepts to conversational systems. However, it remains primarily an empirical contribution rather than advancing new theoretical insights about memory or cognition. Future work could explore deeper theoretical implications for human-AI interaction.

\section*{Ethical Considerations}
Research on episodic memory systems for conversational AI merits thoughtful consideration of privacy aspects, as these systems retain and process user information across multiple sessions. PREMem's structured approach to memory representation offers opportunities for enhanced transparency, potentially enabling clearer user controls over what information is stored. In real-world applications, implementing appropriate data management options would allow users to understand and guide their personalized experience.

The cross-session reasoning capabilities in our approach warrant attention to potential biases and inference accuracy. Our categorization helps distinguish between what users explicitly stated and what the system infers, but misinterpretations can still occur. Future research should develop confidence indicators for memory-based responses and create mechanisms for users to correct the system when it makes inappropriate connections between separate conversations, helping prevent potential misunderstandings from persisting across interactions.

\bibliography{custom}

\input{appendix}
\end{document}

%% file: table/data_category_stats.tex
\begin{table}[ht]
\centering
\resizebox{0.75\columnwidth}{!}{
\begin{tabular}{llc}
\toprule
\textbf{Dataset} & \textbf{Category} & \textbf{\# Questions}\\
\midrule
\multirow{4}{*}{LoCoMo} & single-hop & 1,123 (56.5\%) \\
 & multi-hop & 321 (16.1\%) \\
 & temporal-reasoning & 96 (4.8\%) \\
 & adversarial & 446 (22.4\%) \\
\hline
\addlinespace[0.3em]
\multirow{5}{*}{LongMemEval} & single-hop & 150 (30.0\%) \\
 & multi-hop & 121 (24.2\%) \\
 & temporal-reasoning & 127 (25.4\%) \\
 & adversarial & 30 (6.0\%) \\
 & knowledge-update & 72 (14.4\%) \\
\bottomrule
\end{tabular}
}
\caption{Statistics of dataset category.}
\label{tab:dataset-stats}
\end{table}

%% file: table/main_table_rank.tex
\begingroup
\begin{table*}[!th]
\centering
\resizebox{\textwidth}{!}{%
\renewcommand{\arraystretch}{0.9}     
\fontsize{6}{7}\selectfont
\setlength\tabcolsep{3pt}
\begin{tabular}{
  cc   
  c    
  >{\columncolor{gray!20}}c >{\columncolor{gray!20}}c   
  cc   
  cc   
  cc   
  cc   
  c    
  >{\columncolor{gray!20}}c >{\columncolor{gray!20}}c   
  cc   
  cc   
  cc   
  c    
}
\toprule
\multicolumn{2}{c}{\multirow{3}{*}{\makecell[c]{\textbf{Model}}}}
  & \multirow{3}{*}{\makecell[c]{\textbf{Method}}}
  & \multicolumn{11}{c}{\textbf{LongMemEval}}
  & \multicolumn{9}{c}{\textbf{LoCoMo}} \\
\cmidrule(lr){4-14}\cmidrule(lr){15-23}
  & &
  & \multicolumn{2}{c}{\textbf{Total}}
  & \multicolumn{2}{c}{Single-hop}
  & \multicolumn{2}{c}{Multi-hop}
  & \multicolumn{2}{c}{Temporal}
  & \multicolumn{2}{c}{Knowledge}
  & Adv
  & \multicolumn{2}{c}{\textbf{Total}}
  & \multicolumn{2}{c}{Single-hop}
  & \multicolumn{2}{c}{Multi-hop}
  & \multicolumn{2}{c}{Temporal}
  & Adv \\
\cmidrule(lr){4-5}\cmidrule(lr){6-7}
\cmidrule(lr){8-9}\cmidrule(lr){10-11}
\cmidrule(lr){12-13}\cmidrule(lr){14-14}
\cmidrule(lr){15-16}\cmidrule(lr){17-18}
\cmidrule(lr){19-20}\cmidrule(lr){21-22}
\cmidrule(lr){23-23}
  & &
  & \multicolumn{1}{c}{\textbf{LLM}} & \multicolumn{1}{c}{\textbf{R1}}
  & LLM & R1
  & LLM & R1
  & LLM & R1
  & LLM & R1
  & Acc
  & \multicolumn{1}{c}{\textbf{LLM}} & \multicolumn{1}{c}{\textbf{R1}}
  & LLM & R1
  & LLM & R1
  & LLM & R1
  & Acc \\
\midrule
\multirow{12}{*}{\rotatebox{90}{Qwen2.5}}
  & \multirow{6}{*}{14B}
    & Turn        & 39.7 & 25.3 & 59.4 & 42.6 & 28.3 & 9.0  & 19.5 & 23.5 & 42.5 & 23.5 & \underline{73.3}& 61.6 & 28.9 & \textbf{74.6} & 42.7 & \underline{49.5} & 15.9 & 47.9 & 14.3 & 46.9 \\
  &             & Session     & 29.0 & 19.3 & 53.1 & 37.2 & 16.7 & 5.7  & 13.2 & 17.1 & 15.8 & 9.5 & 66.7& 54.5 & 25.7 & 57.7 & 36.5 & 38.4 & 12.9 & 42.3 & 12.8 & \underline{67.0} \\
  &             & SeCom       & 37.6 & 24.5 & 60.1 & 42.9 & 26.0 & 9.8  & 19.0 & 23.5 & 35.3 & 17.3 & \underline{73.3} & 60.4 & \textbf{31.0} & \underline{72.9} & \textbf{45.8} & 36.4 & 15.1 & 50.2 & \underline{16.3} & 57.4 \\
  &             & HippoRAG-2  & 44.7 & 29.2 & \underline{68.9} & \underline{48.9} & 26.1 & 9.6  & 20.8 & 25.5 & 59.3 & 31.5 & \textbf{75.9} & \underline{61.7} & \underline{30.4} & 69.8 & \underline{44.3} & 45.6 & 15.6 & \underline{54.7} & 14.0 & 64.4 \\
  &             & A-Mem       & \underline{50.3} & \underline{33.0} & \textbf{72.4} & \textbf{53.0} & \underline{34.0} & \underline{14.0} & \underline{30.0} & \underline{26.5} & \underline{63.9} & \underline{38.2} & 66.7 & 43.6 & 30.2 & 52.4 & 34.5 & 44.8 & \underline{29.2} & 38.2 & 14.9 & 23.7 \\
  &             & PREMem      & \textbf{64.7}  & \textbf{40.4}  & 59.5  & 43.3  & \textbf{75.7}  & \textbf{35.0}  & \textbf{48.6}  & \textbf{38.8}  & \textbf{88.3}  & \textbf{56.9}  & 70.0  & \textbf{68.0}  & 29.4  & 69.2  & 38.0  & \textbf{74.1}  & \textbf{30.1}  & \textbf{55.5}  & \textbf{18.0}  & \textbf{69.1} \\

\cmidrule(lr){2-23}
  & \multirow{6}{*}{72B}
& Turn        & 40.6 & 26.2 & 60.8 & 43.4 & 30.2 & 13.8 & 20.6 & 22.8 & 46.5 & 21.8 & 60.0 & \underline{63.7} & 26.3 & \textbf{77.4} & 38.0 & \underline{51.4} & 15.3 & 60.2 & \textbf{17.7} & 47.3 \\
  &            & Session     & 30.9 & 20.8 & 54.8 & 38.4 & 20.0 & 10.8 & 13.0 & 17.1 & 17.7 &  9.1 & \textbf{73.3} & 54.2 & 23.9 & 60.4 & 33.6 & 40.3 & 12.4 & 54.4 & 15.8 & 53.4 \\
  &            & SeCom       & 39.4 & 25.3 & 56.2 & 40.9 & 31.4 & 14.9 & 22.4 & 21.4 & 43.9 & 22.2 & 56.7 & 58.5 & 28.0 & 72.7 & \underline{41.4} & 34.0 & 12.6 & 54.4 & \textbf{17.7} & 47.7 \\
  &            & HippoRAG-2  & 45.9 & 29.8 & \underline{70.6} & \underline{49.9} & 29.6 & 11.8 & 21.8 & 24.6 & 57.1 & 32.7 & \underline{69.0} & 61.6 & \underline{30.4} & 72.1 & \textbf{44.4} & 44.6 & 16.0 & \textbf{62.3} & 16.6 & \underline{57.0} \\
  &            & A-Mem       & \underline{53.6} & \underline{36.2} & \textbf{73.0} & \textbf{55.4} & \underline{38.1} & \underline{16.5} & \underline{39.1} & \underline{31.6} & \underline{66.2} & \underline{41.2} & 60.0 & 45.6 & \textbf{31.7} & 54.6 & 36.2 & 49.8 & \textbf{32.7} & 42.6 & 16.2 & 21.5 \\
  &            & PREMem      & \textbf{67.5} & \textbf{45.4} & 66.6 & 47.1 & \textbf{79.6} & \textbf{45.1} & \textbf{51.6} & \textbf{43.1} & \textbf{85.9} & \textbf{58.8} & 56.7 & \textbf{71.0} & 27.0 & \underline{73.0} & 33.7 & \textbf{76.7} & \underline{30.0} & \underline{61.8} & \underline{17.1} & \textbf{68.8} \\
\midrule
\multirow{12}{*}{\rotatebox{90}{gemma-3}}
  & \multirow{6}{*}{12B}
& Turn        & 36.6  & 22.4  & 55.6  & 40.2  & 24.2  & 9.1   & 22.2  & 17.4  & 46.4  & 22.3  & 40.0  & \underline{45.5} & 27.0  & \underline{65.3} & 42.0  & 40.1  & 13.6  & 28.8  & 6.7   & 3.8   \\
 & & Session     & 28.8  & 16.9  & 51.0  & 34.8  & 16.7  & 7.0   & 15.8  & 11.1  & 15.1  & 8.4   & \textbf{70.0} & 38.0  & 24.2  & 51.6  & 37.1  & 33.1  & 12.6  & 22.0  & 6.9   & 11.7  \\
 & & SeCom       & 37.4  & 23.5  & 55.6  & 39.8  & 26.3  & \underline{10.2} & 23.1  & 19.5  & 43.6  & 24.9  & \underline{50.0} & 44.8  & \underline{30.1} & \textbf{65.4} & \textbf{47.5} & 35.7  & 13.0  & 28.1  & \underline{8.0}  & 4.0   \\
 & & HippoRAG-2  & \underline{43.9} & 27.1  & \textbf{66.2} & \underline{48.2} & \underline{30.0} & 8.3   & \underline{24.1} & 21.2  & \underline{58.1} & \underline{31.0} & 48.3  & 44.3  & 29.7  & 61.8  & \underline{45.9} & 38.9  & 15.6  & \underline{29.1} & 6.4   & 8.9   \\
 & & A-Mem       & 39.0  & \underline{28.1} & \underline{65.5} & \textbf{51.1} & 22.2  & \underline{10.2} & 22.8  & \underline{24.3} & 49.0  & 23.8  & 33.3  & 43.9  & \textbf{31.2} & 47.2  & 31.4  & \underline{40.7} & \underline{20.2} & 22.4  & 6.3   & \textbf{43.8} \\
 & & PREMem      & \textbf{57.7} & \textbf{34.4} & 54.3  & 38.4  & \textbf{63.3} & \textbf{27.1} & \textbf{47.3} & \textbf{31.9} & \textbf{86.3} & \textbf{52.2} & 46.7  & \textbf{50.0} & \underline{30.1} & 61.7  & 43.1  & \textbf{63.9} & \textbf{27.4} & \textbf{36.6} & \textbf{11.2} & \underline{15.5} \\\cmidrule(lr){2-23}
  & \multirow{6}{*}{27B}
& Turn       & 38.0 & 23.6 & 56.4 & 41.5 & 25.1 &  8.9 & 21.8 & 20.0 & 44.3 & 22.7 & \underline{66.7} & \underline{49.7} & 27.5 & \textbf{67.7} & 42.6 & 39.5 & 14.1 & \underline{30.6} &  7.7 & 17.3 \\
 & & Session    & 27.6 & 16.8 & 50.6 & 36.2 & 14.3 &  5.1 & 13.5 & 10.0 & 12.5 &  8.3 & \textbf{73.3} & 43.3 & 24.6 & 53.1 & 37.0 & 32.0 & 11.2 & 22.6 &  8.3 & 32.7 \\
 & & SeCom      & 38.9 & 23.2 & 55.4 & 39.2 & 30.1 & 10.4 & 24.0 & 19.6 & 40.8 & 22.8 & 63.3 & 49.1 & 30.2 & \underline{67.5} & \textbf{48.0} & 33.2 & 10.9 & 30.0 & \textbf{10.6} & 19.7 \\
 & & HippoRAG-2 & 43.1 & 27.3 & \underline{65.0} & \underline{47.4} & 28.4 & \underline{12.4} & 22.8 & 21.0 & 56.5 & 28.6 & 63.3 & 49.5 & 30.6 & 64.7 & \underline{46.8} & 37.7 & 16.4 & 28.8 &  7.1 & 26.2 \\
 & & A-Mem      & \underline{45.3} & \underline{31.9} & \textbf{66.2} & \textbf{54.5} & \underline{30.5} & 10.6 & \underline{28.9} & \underline{26.1} & \underline{61.7} & \underline{39.2} & 43.3 & 44.5 & \textbf{32.8} & 48.7 & 32.9 & \underline{43.2} & \textbf{28.7} & 24.2 &  7.2 & \textbf{40.0} \\
 & & PREMem     & \textbf{61.9} & \textbf{39.2} & 52.7 & 39.8 & \textbf{69.6} & \textbf{33.4} & \textbf{51.2} & \textbf{38.5} & \textbf{91.3} & \textbf{60.1} & \underline{66.7} & \textbf{54.6} & \underline{30.6} & 62.5 & 40.3 &\textbf{57.0} &  \underline{25.2} & \textbf{34.8} & \underline{8.8} & \underline{38.3} \\
\midrule
\multirow{12}{*}{\rotatebox{90}{gpt-4.1}}

  & \multirow{6}{*}{mini}
    & Turn        & 39.5 & 25.4 & 62.8 & 43.8 & 27.6 & 11.8 & 20.4 & 21.3 & 45.5 & 23.7 & 46.7
                 & \underline{54.7} & 30.3 & \textbf{74.3} & 45.8 & 50.3 & 17.7 & \underline{49.8} & \underline{17.5} & 10.5 \\
  &            & Session     & 29.7 & 18.0 & 54.4 & 37.3 & 18.4 &  6.2 & 13.2 & 12.8 & 17.6 &  9.0 & \underline{66.7}
                 & 48.1 & 27.5 & 58.5 & 41.3 & 41.4 & 15.4 & 38.2 & 16.9 & 30.3 \\
  &            & SeCom       & 42.3 & 26.8 & 64.3 & 45.2 & 33.5 & 13.6 & 24.8 & 24.2 & 41.6 & 21.2 & 60.0
                 & 53.4 & 33.2 & \underline{74.2} & \textbf{51.6} & 40.1 & 17.0 & 42.0 & 11.7 & 14.1 \\
  &            & HippoRAG-2  & 44.8 & 29.1 & \underline{69.5} & \underline{48.3} & 30.4 & 12.2 & 19.5 & 23.2 & 56.3 & 33.1 & \textbf{72.0}
                 & 54.6 & 34.1 & 70.0 & \underline{50.2} & 52.9 & 25.0 & 42.6 & 17.0 & 23.3 \\
  &            & A-Mem       & \underline{53.9} & \underline{35.5} & \textbf{75.1} & \textbf{57.4} & \underline{41.6} & \underline{16.5} & \underline{34.1} & \underline{27.9} & \underline{67.4} & \underline{39.0} & \underline{66.7}
                 & 52.7 & \textbf{37.0} & 56.1 & 36.8 & \underline{61.1} & \textbf{38.0} & 38.5 & 11.0 & \underline{42.5} \\
  &            & PREMem      & \textbf{67.6} & \textbf{43.2} & 56.4 & 40.5 & \textbf{76.5} & \textbf{41.7} & \textbf{62.4} & \textbf{44.3} & \textbf{88.6} & \textbf{60.4} & 63.3
                 & \textbf{64.9} & \underline{34.5} & 69.4 & 46.7 & \textbf{77.9} & \underline{36.9} & \textbf{50.3} & \textbf{18.2} & \textbf{48.9} \\
\cmidrule(lr){2-23}
  & \multirow{6}{*}{base}
    & Turn        & 40.7 & 25.2 & 61.8 & 43.6 & 25.8 &  9.6 & 24.1 & 22.5 & 47.5 & 23.9 & 56.7
                 & 57.1 & 31.3 & \textbf{76.3} & 45.9 & 54.7 & 21.7 & \underline{53.4} & 20.5 & 12.8 \\
  &            & Session     & 30.3 & 18.3 & 54.9 & 37.6 & 20.6 &  9.0 & 10.3 & 11.0 & 14.8 &  9.2 & \textbf{76.7}
                 & 50.1 & 27.9 & 59.6 & 40.6 & 42.2 & 17.0 & 49.1 & \underline{20.7} & \underline{34.1} \\
  &            & SeCom       & 42.0 & 26.2 & 63.3 & 44.2 & 32.9 & 13.2 & 20.3 & 21.4 & 49.0 & 24.2 & 60.0
                 & 56.7 & \underline{35.0} & \underline{76.2} & \textbf{52.9} & 42.5 & 19.0 & 50.6 & 19.7 & 20.9 \\
  &            & HippoRAG-2  & 45.2 & 29.2 & \underline{70.3} & \underline{50.5} & 28.2 & 11.3 & 19.1 & 21.6 & 58.3 & 34.6 & \textbf{76.7}
                 & \underline{57.3} & 34.0 & 71.6 & \underline{49.8} & 49.4 & 22.4 & \textbf{54.9} & \textbf{22.6} & 30.9 \\
  &            & A-Mem       & \underline{55.9} & \underline{37.5} & \textbf{78.0} & \textbf{61.3} & \underline{41.4} & \underline{17.0} & \underline{37.8} & \underline{30.9} & \underline{64.7} & \underline{39.2} & 66.7
                 & 49.5 & 34.7 & 55.6 & 36.6 & \underline{58.6} & \textbf{39.8} & 39.6 & 11.5 & 30.9 \\
  &            & PREMem      & \textbf{71.4} & \textbf{44.6} & 58.5 & 40.9 & \textbf{83.5} & \textbf{44.0} & \textbf{64.4} & \textbf{44.8} & \textbf{93.7} & \textbf{64.8} & \underline{73.3}
                 & \textbf{67.7} & \textbf{35.9} & 71.5 & 48.5 & \textbf{76.0} & \underline{36.4} & 50.2 & 19.4 & \textbf{57.4} \\
\bottomrule
\end{tabular}%
}
\caption{Performance comparison across different model sizes and memory frameworks. Results show LLM-judge scores (LLM), ROUGE-1 (R1), and adversarial accuracy (Acc). Highest scores in \textbf{bold} and second highest \underline{underlined}. Additional metrics (BLEU-1, ROUGE-L, METEOR, BERTScore) available in Appendix \ref{apdx:complementary_results}.}
\label{tab:main_results}
\end{table*}
\endgroup

%% file: table/small_model_1_3.tex
\begin{table}[h]
\resizebox{\columnwidth}{!}{%
\renewcommand{\arraystretch}{0.9}     
\fontsize{6}{7}\selectfont
\setlength\tabcolsep{4pt}
\begin{tabular}{
  c    
  c    
  c    
  c    
}
\toprule
\textbf{Method} & \textbf{Model} & \textbf{LongMemEval} & \textbf{LoCoMo} \\
\midrule
Turn       & \multirow{5}{*}{Qwen2.5 72B}   & 40.6          & \textbf{63.7}          \\
Session    &                                & 30.9          & 54.2          \\
SeCom      &                                & 39.4          & 58.5          \\
HippoRAG-2 &                                & 45.9          & \underline{61.6}          \\
A-Mem      &                                & \textbf{53.6}          & 45.6          \\ \cmidrule(lr){1-4}
PREMem     &  Qwen2.5 3B                    & \underline{50.8}          & 53.8          \\
\midrule
Turn       & \multirow{5}{*}{gemma-3 27B}  & 38.0          & \underline{49.7}          \\
Session    &                               & 27.6          & 33.3          \\
SeCom      &                               & 38.9          & 49.1          \\
HippoRAG-2 &                               & 43.1          & 49.5          \\
A-Mem      &                               & \underline{45.3}          & 44.5          \\ \cmidrule(lr){1-4}
PREMem     & gemma-3 4B                    & \textbf{53.4}          & \textbf{50.1}          \\
\midrule
Turn       & \multirow{5}{*}{gpt-4.1}      & 40.7          & 57.1          \\
Session    &                               & 30.3          & 50.1          \\
SeCom      &                               & 42.0          & 56.7          \\
HippoRAG-2 &                               & 45.2          & \underline{57.3}          \\
A-Mem      &                               & \underline{55.9}          & 49.5          \\ \cmidrule(lr){1-4}
PREMem     & gpt-4.1 nano                  &\textbf{ 58.7 }         & \textbf{58.8}          \\
\bottomrule
\end{tabular}%
}
\caption{Small models with PREMem vs. larger models with baselines (LLM-as-a-judge scores).}
\label{tab:small}
\end{table}

%% file: table/Ablation_table.tex
\begingroup
\setlength{\tabcolsep}{6pt}
\renewcommand{\arraystretch}{0.7}
\begin{table*}[b]
\centering
\scriptsize 
\resizebox{\textwidth}{!}{%
\fontsize{5.5pt}{6pt}\selectfont 
\begin{tabular}{
  l
  *{6}{c}
  *{6}{c}
}
\toprule
\addlinespace[1pt]
\multirow{3}{*}{\textbf{Method}} 
& \multicolumn{6}{c}{\textbf{LLM}} & \multicolumn{6}{c}{\textbf{R1}} \\ \addlinespace[-1pt]
\cmidrule(lr){2-7} \cmidrule(lr){8-13}
\addlinespace[-1.5pt]
&  \multicolumn{2}{c}{Qwen2.5} & \multicolumn{2}{c}{gemma-3} & \multicolumn{2}{c}{gpt-4.1}
&  \multicolumn{2}{c}{Qwen2.5} & \multicolumn{2}{c}{gemma-3} & \multicolumn{2}{c}{gpt-4.1} \\
\addlinespace[-1pt]
\cmidrule(lr){2-3} \cmidrule(lr){4-5} \cmidrule(lr){6-7} \cmidrule(lr){8-9} \cmidrule(lr){10-11} \cmidrule(lr){12-13}
\addlinespace[-1pt]
& 14B & 72B & 12B & 27B & mini & base
& 14B & 72B & 12B & 27B & mini & base \\
\addlinespace[-1pt]
\midrule
\addlinespace[1pt]
\multicolumn{13}{c}{\textbf{LongMemEval}} \\
\addlinespace[-1pt]
\midrule
\addlinespace[1pt]
PREMem
& 64.7 & 67.5 & 57.7 & 61.9 & 67.6 & 71.4
& 40.4 & 45.4 & 34.4 & 39.2 & 43.2 & 44.6 \\
\cdashline{1-13}[0.4pt/1pt]
\addlinespace[1pt]
w/o Step 2
& \makecell{65.0 \\[-2pt] {\fontsize{4pt}{3.5pt}\selectfont(+0.5\%)}} & \makecell{69.8 \\ [-2pt] {\fontsize{4pt}{3.5pt}\selectfont(+3.5\%)}} & \makecell{57.4 \\ [-2pt] {\fontsize{4pt}{3.5pt}\selectfont(-0.6\%)}} & \makecell{59.9 \\ [-2pt] {\fontsize{4pt}{3.5pt}\selectfont(-3.2\%)}} & \makecell{67.9 \\ [-2pt] {\fontsize{4pt}{3.5pt}\selectfont(+0.4\%)}} & \makecell{69.8 \\ [-2pt] {\fontsize{4pt}{3.5pt}\selectfont(-2.2\%)}}
& \makecell{39.6 \\ [-2pt] {\fontsize{4pt}{3.5pt}\selectfont(-2.1\%)}} & \makecell{45.2 \\ [-2pt] {\fontsize{4pt}{3.5pt}\selectfont(-0.3\%)}} & \makecell{\underline{32.5} \\ [-2pt] {\fontsize{4pt}{3.5pt}\selectfont(-5.5\%)}} & \makecell{38.1 \\ [-2pt] {\fontsize{4pt}{3.5pt}\selectfont(-2.7\%)}} & \makecell{43.3 \\ [-2pt] {\fontsize{4pt}{3.5pt}\selectfont(+0.3\%)}} & \makecell{43.4 \\ [-2pt] {\fontsize{4pt}{3.5pt}\selectfont(-2.7\%)}} \\
\cdashline{1-13}[0.4pt/1pt]
\addlinespace[1pt]
w/o Step 1
& \makecell{\textbf{31.2} \\ [-2pt] {\fontsize{4pt}{3.5pt}\selectfont(-51.8\%)}} & \makecell{\textbf{35.9} \\ [-2pt] {\fontsize{4pt}{3.5pt}\selectfont(-46.8\%)}} & \makecell{\textbf{23.5} \\ [-2pt] {\fontsize{4pt}{3.5pt}\selectfont(-59.3\%)}} & \makecell{\textbf{24.1} \\ [-2pt] {\fontsize{4pt}{3.5pt}\selectfont(-61.0\%)}} & \makecell{\textbf{31.9} \\ [-2pt] {\fontsize{4pt}{3.5pt}\selectfont(-52.8\%)}} & \makecell{\textbf{34.2} \\ [-2pt] {\fontsize{4pt}{3.5pt}\selectfont(-52.1\%)}}
& \makecell{\textbf{17.2} \\ [-2pt] {\fontsize{4pt}{3.5pt}\selectfont(-57.4\%)}} & \makecell{\textbf{18.4} \\ [-2pt] {\fontsize{4pt}{3.5pt}\selectfont(-59.5\%)}} & \makecell{\textbf{11.1} \\ [-2pt] {\fontsize{4pt}{3.5pt}\selectfont(-67.8\%)}} & \makecell{\textbf{12.2} \\ [-2pt] {\fontsize{4pt}{3.5pt}\selectfont(-69.0\%)}} & \makecell{\textbf{15.8} \\ [-2pt] {\fontsize{4pt}{3.5pt}\selectfont(-63.5\%)}} & \makecell{\textbf{17.2} \\ [-2pt] {\fontsize{4pt}{3.5pt}\selectfont(-61.5\%)}} \\
\cdashline{1-13}[0.4pt/1pt]
\addlinespace[1pt]
w/o Step 1 Categories
& \makecell{64.3 \\ [-2pt] {\fontsize{4pt}{3.5pt}\selectfont(-0.7\%)}} & \makecell{68.9 \\ [-2pt] {\fontsize{4pt}{3.5pt}\selectfont(+2.0\%)}} & \makecell{56.3 \\ [-2pt] {\fontsize{4pt}{3.5pt}\selectfont(-2.4\%)}} & \makecell{59.9 \\ [-2pt] {\fontsize{4pt}{3.5pt}\selectfont(-3.2\%)}} & \makecell{66.7 \\ [-2pt] {\fontsize{4pt}{3.5pt}\selectfont(-1.3\%)}} & \makecell{69.6 \\ [-2pt] {\fontsize{4pt}{3.5pt}\selectfont(-2.4\%)}}
& \makecell{40.3 \\ [-2pt] {\fontsize{4pt}{3.5pt}\selectfont(-0.3\%)}} & \makecell{45.7 \\ [-2pt] {\fontsize{4pt}{3.5pt}\selectfont(+0.8\%)}} & \makecell{33.0 \\ [-2pt] {\fontsize{4pt}{3.5pt}\selectfont(-4.1\%)}} & \makecell{38.1 \\ [-2pt] {\fontsize{4pt}{3.5pt}\selectfont(-2.7\%)}} & \makecell{41.9 \\ [-2pt] {\fontsize{4pt}{3.5pt}\selectfont(-2.9\%)}} & \makecell{42.9 \\ [-2pt] {\fontsize{4pt}{3.5pt}\selectfont(-3.7\%)}} \\
\cdashline{1-13}[0.4pt/1pt]
\addlinespace[1pt]
w/o Temporal Reasoning
& \makecell{63.7 \\ [-2pt] {\fontsize{4pt}{3.5pt}\selectfont(-1.6\%)}} & \makecell{68.4 \\ [-2pt] {\fontsize{4pt}{3.5pt}\selectfont(+1.3\%)}} & \makecell{56.0 \\ [-2pt] {\fontsize{4pt}{3.5pt}\selectfont(-3.0\%)}} & \makecell{\underline{58.5} \\ [-2pt] {\fontsize{4pt}{3.5pt}\selectfont(-5.4\%)}} & \makecell{66.2 \\ [-2pt] {\fontsize{4pt}{3.5pt}\selectfont(-2.1\%)}} & \makecell{69.0 \\ [-2pt] {\fontsize{4pt}{3.5pt}\selectfont(-3.4\%)}}
& \makecell{39.1 \\ [-2pt] {\fontsize{4pt}{3.5pt}\selectfont(-3.2\%)}} & \makecell{44.8 \\ [-2pt] {\fontsize{4pt}{3.5pt}\selectfont(-1.2\%)}} & \makecell{\textbf{30.8} \\ [-2pt] {\fontsize{4pt}{3.5pt}\selectfont(-10.6\%)}} & \makecell{\underline{36.5} \\ [-2pt] {\fontsize{4pt}{3.5pt}\selectfont(-6.8\%)}} & \makecell{42.9 \\ [-2pt] {\fontsize{4pt}{3.5pt}\selectfont(-0.6\%)}} & \makecell{43.9 \\ [-2pt] {\fontsize{4pt}{3.5pt}\selectfont(-1.6\%)}} \\
\addlinespace[-1pt]
\midrule
\addlinespace[1pt]
\multicolumn{13}{c}{\textbf{LoCoMo}} \\
\addlinespace[-1pt]
\midrule
\addlinespace[1pt]
PREMem
& 68.0 & 71.0 & 50.0 & 54.6 & 64.9 & 67.7
& 29.4 & 27.0 & 30.1 & 30.6 & 34.5 & 35.9 \\
\cdashline{1-13}[0.4pt/1pt]
\addlinespace[1pt]
w/o Step 2
& \makecell{\underline{64.4} \\ [-2pt] {\fontsize{4pt}{3.5pt}\selectfont(-5.3\%)}} & \makecell{68.2 \\ [-2pt] {\fontsize{4pt}{3.5pt}\selectfont(-3.8\%)}} & \makecell{\underline{47.3} \\ [-2pt] {\fontsize{4pt}{3.5pt}\selectfont(-5.4\%)}} & \makecell{52.8 \\ [-2pt] {\fontsize{4pt}{3.5pt}\selectfont(-3.2\%)}} & \makecell{\underline{61.4} \\ [-2pt] {\fontsize{4pt}{3.5pt}\selectfont(-5.4\%)}} & \makecell{64.7 \\ [-2pt] {\fontsize{4pt}{3.5pt}\selectfont(-4.5\%)}}
& \makecell{29.6 \\ [-2pt] {\fontsize{4pt}{3.5pt}\selectfont(+0.6\%)}} & \makecell{28.6 \\ [-2pt] {\fontsize{4pt}{3.5pt}\selectfont(+5.6\%)}} & \makecell{\underline{28.3} \\ [-2pt] {\fontsize{4pt}{3.5pt}\selectfont(-6.0\%)}} & \makecell{\underline{28.5} \\ [-2pt] {\fontsize{4pt}{3.5pt}\selectfont(-7.0\%)}} & \makecell{33.2 \\ [-2pt] {\fontsize{4pt}{3.5pt}\selectfont(-3.6\%)}} & \makecell{34.2 \\ [-2pt] {\fontsize{4pt}{3.5pt}\selectfont(-4.8\%)}} \\
\cdashline{1-13}[0.4pt/1pt]
\addlinespace[1pt]
w/o Step 1
& \makecell{\textbf{44.5} \\ [-2pt] {\fontsize{4pt}{3.5pt}\selectfont(-34.6\%)}} & \makecell{\textbf{47.8} \\ [-2pt] {\fontsize{4pt}{3.5pt}\selectfont(-32.7\%)}} & \makecell{\textbf{32.3} \\ [-2pt] {\fontsize{4pt}{3.5pt}\selectfont(-35.4\%)}} & \makecell{\textbf{33.9} \\ [-2pt] {\fontsize{4pt}{3.5pt}\selectfont(-37.8\%)}} & \makecell{\textbf{41.9} \\ [-2pt] {\fontsize{4pt}{3.5pt}\selectfont(-35.4\%)}} & \makecell{\textbf{44.1} \\ [-2pt] {\fontsize{4pt}{3.5pt}\selectfont(-34.9\%)}}
& \makecell{\textbf{14.7} \\ [-2pt] {\fontsize{4pt}{3.5pt}\selectfont(-49.9\%)}} & \makecell{\textbf{14.6} \\ [-2pt] {\fontsize{4pt}{3.5pt}\selectfont(-45.9\%)}} & \makecell{\textbf{13.1} \\ [-2pt] {\fontsize{4pt}{3.5pt}\selectfont(-56.4\%)}} & \makecell{\textbf{13.2} \\ [-2pt] {\fontsize{4pt}{3.5pt}\selectfont(-56.7\%)}} & \makecell{\textbf{16.5} \\ [-2pt] {\fontsize{4pt}{3.5pt}\selectfont(-52.1\%)}} & \makecell{\textbf{17.9} \\ [-2pt] {\fontsize{4pt}{3.5pt}\selectfont(-50.1\%)}} \\
\cdashline{1-13}[0.4pt/1pt]
\addlinespace[1pt]
w/o Step 1 Categories
& \makecell{65.7 \\ [-2pt] {\fontsize{4pt}{3.5pt}\selectfont(-3.4\%)}} & \makecell{68.1 \\ [-2pt] {\fontsize{4pt}{3.5pt}\selectfont(-4.1\%)}} & \makecell{49.1 \\ [-2pt] {\fontsize{4pt}{3.5pt}\selectfont(-1.8\%)}} & \makecell{52.4 \\ [-2pt] {\fontsize{4pt}{3.5pt}\selectfont(-4.0\%)}} & \makecell{\underline{60.8} \\ [-2pt] {\fontsize{4pt}{3.5pt}\selectfont(-6.2\%)}} & \makecell{\underline{63.5} \\ [-2pt] {\fontsize{4pt}{3.5pt}\selectfont(-6.3\%)}}
& \makecell{\underline{27.9} \\ [-2pt] {\fontsize{4pt}{3.5pt}\selectfont(-5.3\%)}} & \makecell{26.3 \\ [-2pt] {\fontsize{4pt}{3.5pt}\selectfont(-2.9\%)}} & \makecell{\underline{27.5} \\ [-2pt] {\fontsize{4pt}{3.5pt}\selectfont(-8.7\%)}} & \makecell{\underline{27.9} \\ [-2pt] {\fontsize{4pt}{3.5pt}\selectfont(-8.9\%)}} & \makecell{\underline{32.0} \\ [-2pt] {\fontsize{4pt}{3.5pt}\selectfont(-7.2\%)}} & \makecell{\underline{33.9} \\ [-2pt] {\fontsize{4pt}{3.5pt}\selectfont(-5.6\%)}} \\
\cdashline{1-13}[0.4pt/1pt]
\addlinespace[1pt]
w/o Temporal Reasoning
& \makecell{\underline{64.2} \\ [-2pt] {\fontsize{4pt}{3.5pt}\selectfont(-5.7\%)}} & \makecell{\underline{65.8} \\ [-2pt] {\fontsize{4pt}{3.5pt}\selectfont(-7.3\%)}} & \makecell{47.8 \\ [-2pt] {\fontsize{4pt}{3.5pt}\selectfont(-4.3\%)}} & \makecell{52.8 \\ [-2pt] {\fontsize{4pt}{3.5pt}\selectfont(-3.2\%)}} & \makecell{\underline{60.9} \\ [-2pt] {\fontsize{4pt}{3.5pt}\selectfont(-6.1\%)}} & \makecell{\underline{62.6} \\ [-2pt] {\fontsize{4pt}{3.5pt}\selectfont(-7.6\%)}}
& \makecell{\underline{27.4} \\ [-2pt] {\fontsize{4pt}{3.5pt}\selectfont(-6.9\%)}} & \makecell{26.4 \\ [-2pt] {\fontsize{4pt}{3.5pt}\selectfont(-2.5\%)}} & \makecell{\textbf{25.1} \\ [-2pt] {\fontsize{4pt}{3.5pt}\selectfont(-16.4\%)}} & \makecell{\textbf{26.8} \\ [-2pt] {\fontsize{4pt}{3.5pt}\selectfont(-12.6\%)}} & \makecell{\underline{31.1} \\ [-2pt] {\fontsize{4pt}{3.5pt}\selectfont(-9.9\%)}} & \makecell{\underline{32.7} \\ [-2pt] {\fontsize{4pt}{3.5pt}\selectfont(-9.1\%)}} \\
\addlinespace[-1pt]
\bottomrule
\end{tabular}%
}
\caption{Ablation study of PREMem Components. \textbf{Bold}: >10\% drop, \underline{underlined}: 5-10\% drop from PREMem.}
\label{tab:ablation}
\end{table*}
\endgroup

%% file: table/token_budget.tex
\begingroup

\setlength{\tabcolsep}{2pt}

\renewcommand{\arraystretch}{0.9}
\setlength{\aboverulesep}{0.2pt}
\setlength{\belowrulesep}{0.2pt}
\setlength{\extrarowheight}{1pt}

\begin{table}[ht]
\centering
\resizebox{\columnwidth}{!}{%
\renewcommand{\arraystretch}{0.9}
\fontsize{6}{7}\selectfont
\setlength\tabcolsep{3pt}
\begin{tabular}{
  cc
  ccccccc
}
\toprule
\multirow{3}{*}{\textbf{Method}} & \multirow{3}{*}{\makecell{\textbf{Token}\\\textbf{budget}}} 
& \multicolumn{6}{c}{\textbf{Model}} \\
\cmidrule(lr){3-8}
& & \multicolumn{2}{c}{Qwen2.5} & \multicolumn{2}{c}{gemma-3} & \multicolumn{2}{c}{gpt-4.1} \\
\cmidrule(lr){3-8}
& & 14B & 72B & 12B & 27B & mini & base \\
\midrule
\multicolumn{8}{c}{\textbf{LongMemEval}} \\
\midrule

\multirow{3}{*}{SeCom} &
1024
& 35.8 & 37.2 & 33.4 & 33.0 & 37.0 & 35.5 \\
&2048
& 37.6 & 39.4 & 37.4 & 38.9 & 42.3 & 42.0 \\
&4096
& \textbf{42.8} & \textbf{44.4} & \textbf{38.8} & \textbf{40.4} & \textbf{44.3} & \textbf{44.4} \\
\cmidrule(lr){1-8}
\multirow{3}{*}{A-Mem} &
1024
& 44.4 & 48.6 & 36.4 & 41.0 & 49.2 & 49.4 \\
&2048
& 50.3 & 53.6 & 39.0 & 45.3 & 53.9 & 55.9 \\
&4096
& \textbf{54.8} & \textbf{58.5} & \textbf{44.9} & \textbf{50.6} & \textbf{61.3} & \textbf{62.0} \\
\cmidrule(lr){1-8}
\multirow{3}{*}{HippoRAG-2} &
1024
& 41.5 & 41.5 & 38.5 & 38.3 & 40.8 & 40.2 \\
&2048
& 44.7 & 45.9 & 43.9 & 43.1 & 44.8 & 45.2 \\
&4096
& \textbf{57.5} & \textbf{57.4} & \textbf{51.3} & \textbf{53.5} & \textbf{61.0} & \textbf{61.7} \\
\cmidrule(lr){1-8}
\multirow{3}{*}{PREMem} & 
1024
& \textbf{66.4} & \textbf{67.6} & \textbf{58.9} & \textbf{63.0} & \textbf{68.7} & 70.2 \\
&2048
& 64.7 & 67.5 & 57.7 & 61.9 & 67.6 & 71.4 \\
&4096
& 62.2 & 66.9 & 55.7 & 60.5 & 67.2 & \textbf{71.8} \\
\midrule
\multicolumn{8}{c}{\textbf{LoCoMo}} \\
\midrule

\multirow{3}{*}{SeCom} & 1024
& 57.0 & 60.5 & 42.3 & 47.9 & 51.5 & 54.2 \\
& 2048
& 60.4 & 58.5 & 44.8 & 49.1 & 53.4 & 56.7 \\
& 4096
& \textbf{63.4} & \textbf{63.9} & \textbf{46.0} & \textbf{50.1} & \textbf{54.6} & \textbf{57.3} \\
\cmidrule(lr){1-8}
\multirow{3}{*}{A-Mem} & 1024
& 42.9 & \textbf{45.9} & 43.8 & \textbf{44.5} & 52.9 & \textbf{51.5} \\
& 2048
& \textbf{43.6} & 45.6 & \textbf{43.9} & \textbf{44.5} & 52.7 & 49.5 \\
& 4096
& 43.5 & 45.1 & 43.8 & 44.3 & \textbf{53.3} & 50.2 \\
\cmidrule(lr){1-8}
\multirow{3}{*}{HippoRAG-2} & 1024
& 56.0 & 55.7 & 40.8 & 46.5 & 49.7 & 53.1 \\
& 2048
& 61.7 & 61.6 & 44.3 & 49.5 & 54.6 & 57.3 \\
& 4096
& \textbf{64.1} & \textbf{65.3} & \textbf{47.0} & \textbf{51.0} & \textbf{56.2} & \textbf{59.5} \\
\cmidrule(lr){1-8}
\multirow{3}{*}{PREMem} & 1024
& 63.7 & 65.6 & 48.1 & 52.7 & 64.6 & 67.3 \\
& 2048
& \textbf{68.0} & \textbf{71.0} & \textbf{50.0} & \textbf{54.6} & \textbf{64.9} & \textbf{67.7} \\
& 4096
& 67.0 & 68.7 & 47.2 & 53.3 & 64.8 & 67.1 \\
\bottomrule
\end{tabular}%
}
\caption{Performance across token budgets. \textbf{Bold} indicates the highest score in each range.}

\label{tab:token-budget}
\end{table}
\endgroup

%% file: appendix.tex
\appendix

\section{LLM Prompt}
\label{apdx:llm_prompt}
We include all prompts required to run and evaluate PREMem. In these figures, placeholders (denoted by \{\{\$variable\}\}) indicate positions where specific content is dynamically inserted during execution. For Step 1, refer to Figure \ref{fig:step1-prompt}, for Step 2, refer to Figure \ref{fig:step2-prompt}. The response generation prompts for LongMemEval and LoCoMo are provided in Figure \ref{fig:longmemeval-answer-generation-prompt} and Figure \ref{fig:locomo-answer-generation-prompt}, respectively. The LLM-as-a-judge evaluation prompt is shown in Figure \ref{fig:llm-as-a-judge}.

\input{prompts/step_1}

\input{prompts/step_2}

\begin{figure}[h]
\centering
\begin{tcolorbox}
\begin{spacing}{1.0}
\small{
Based on the context, write an answer in the form of a short phrase for the following question. Answer with exact words from the context whenever possible.\\
\\
Context:\\
\{\{\$context\}\}\\
\\
Question: \{\{\$question\}\}\\
\\
Short Answer:}
\end{spacing}\noindent
\end{tcolorbox}
\caption{LoCoMo answer generation prompt.}
\label{fig:locomo-answer-generation-prompt}
\end{figure}

\begin{figure}[t]
\centering
\begin{tcolorbox}
\begin{spacing}{1.0}
\small{
You are an intelligent assistant designed to provide concise, accurate answers based on given context. Your task is to analyze the provided information and respond to a specific question with a few words or a short phrase.\\
\\
Here is the context you should use to inform your answer:\\
\\
\{\{\$context\}\}\\
\\
Now, please consider the following question:\\
\\
\{\{\$question\}\}\\
\\
Instructions:\\
1. Carefully read and analyze the provided context.\\
2. Consider the question in relation to the context.\\
3. Formulate a concise answer based solely on the information given in the context.\\
4. Respond with a short phrase only. Do not use a full sentence.\\
5. Do not include any explanations, reasoning, or greetings in your response.\\
6. Ensure your answer is directly relevant to the question asked.\\
\\
Your response should provide only the essential information in a brief phrase.\\
\\
Answer:
}
\end{spacing}\noindent
\end{tcolorbox}
\caption{LongMemEval answer generation prompt.}
\label{fig:longmemeval-answer-generation-prompt}
\end{figure}


\section{Cognitive Scientific Foundation for Memory Evolution Patterns}
\label{apdx:schema}

In cognitive science, schema is a framework (structure) that organizes an individual's experiences, knowledge, and information, as well as the way they are stored in his memory.
The schema stores one's experiences, knowledge and information as its memory, and it is developed by assimilating and accommodating the information \cite{piaget1952origins}.
When new information aligns with an existing schema, the schema assimilation occurs.
Conversely, misaligned information requires schema updates to incorporate new data.

A seminal work in schema theory \cite{rumelhart1976accretion} introduced three modes of learning: accretion, tuning, and restructuring.
This work has become the foundation of understanding how existing knowledge structures—known as schemata—are transformed whenever new information is encountered.
In particular, accretion means adding new information to an existing schema without altering its structure.
Tuning refines the existing schema, making it more efficient or accurate.
Restructuring, on the other hand, involves a more fundamental change in the schema’s structure.
Thus, this work has become the foundation for further investigations on how schemata are modified and reorganized in response to new informaion.

Referring to further schema theory literature \cite{chi1981expertise, bartlett1995remembering, mandler2014stories, rumelhart2017schemata}, we identify six major mechanisms of how a schema modified:
(1) Schema expansion refers to adding a new attribute or feature to an existing schema;
(2) Schema integration occurs when separate, related schemata become connected to form a more cohesive structure;
(3) Schema refinement points to the process of a schema being refined or made more specific based on accumulated details;
(4) Schema reinforcement happens when similar information is repeatedly acquired, strengthening the existing schema;
(5) Schema restructuring completely reorganizes schema structure;
and (6) Schema creation occurs when existing schema structure does not align with a new information, leading to the creation of an entirely new schema.

During a conversation, the individual acquires additional information, and integrates new information into established memory.
When integrating, it is crucial to consider how the new information is related to prior memory.
Referring to cognitive science \cite{anderson2013architecture, bransford1972contextual}, which studies how people perceive and learn from information, and conceptual development \cite{carey1985conceptual, murphy2004big, chi2009three}, which studies how infants learn concepts, we identify five information types-- extension, accumulation, specification, transformation, and connection--each of which causes a different type of modification in the underlying schema.

\paragraph{Extension (Elaboration)} A new information broadens the scope of existing knowledge.
\cite{anderson2013architecture, carey1985conceptual} describe that exposure to information and experience extend the existing knowledge structure, paralleling the process of schema expansion.

\paragraph{Accumulation} The similar type of information accumulates.
Repeated exposure to similar information solidifies an existing framework.
\cite{chi1981expertise, schank2013scripts} demonstrate that repeated encounters with similar information and experience solidify a schema.

\paragraph{Specification} The existing information becomes more detailed and developed more precisely.
New information refines existing knowledge by adding more detailed or precise features, causing schema refinement.
\cite{murphy2004big, keil1979semantic} both claim that knowledge is refined and differentiated as precise and specific information is encountered.

\paragraph{Transformation} The previous information is replaced by new information or fundamentally modified.
New information drives schema restructuring.
According to \cite{chi2009three, rumelhart1976accretion}, schema is reconstructed when the new information does not fit to the existing knowledge significantly.

\paragraph{Connection} The relationship between the information and the causality are revealed.
The connected information promotes existing schemata to be integrated.
\cite{bransford1972contextual, fauconnier2008way} show that connection between information develops individual's reasoning and understanding.

These five types of new information—extension, accumulation, specification, transformation, and connection—are consistent with the schema modification mechanisms: schema expansion, reinforcement, refinement, restructuring, and integration.

\section{Dataset Description and Category Unification}
\label{apdx:dataset}

For consistency, we unify the question types into five categories: \textit{single-hop}, \textit{multi-hop}, \textit{temporal reasoning}, \textit{adversarial}, and \textit{knowledge update} (LongMemEval only). For LoCoMo, we treat all questions originally labeled as \texttt{open-domain-knowledge} as \texttt{single-hop}. The other labels—\texttt{multi-hop}, \texttt{temporal reasoning}, and \texttt{adversarial}—are retained as-is. For LongMemEval, we apply these mappings: Any type containing the word \texttt{single} is mapped to \texttt{single-hop}. All other types are converted by replacing \texttt{session} with \texttt{hop}, aligning them with the \texttt{multi-hop} or \texttt{temporal reasoning} categories. If the question ID ends with \texttt{\_abs}, it is classified as \texttt{adversarial} based on its original designation as an abstention question. Questions related to knowledge revision are assigned to the \texttt{knowledge update} category. 

This unified labeling scheme supports direct comparison across datasets and is used for all category-level evaluations in this work. Datasets are available under CC-BY-NC-4.0 (LoCoMo) and MIT License (LongMemEval).

\onecolumn
\section{Complementary Results}
\label{apdx:complementary_results}

We present the complete scores including metrics that were omitted from the main paper in Table \ref{tab:all}.
\input{table/all_table}

\twocolumn

\section{Implementation Details}
\label{apdx:implementation}

For our implementation, we set the threshold parameter $\theta$ to 0.6 for memory fragment selection. In Steps 1 and 2 of our methodology, we utilized few-shot examples to enhance performance, with the complete prompt templates available in Appendix \ref{apdx:llm_prompt}. To ensure consistent evaluation across experiments, we conducted preference testing to determine which answers were more favorable. Our analysis revealed no statistically significant difference between using GPT-4o and GPT-4.1-mini as judges, leading us to select GPT-4.1-mini as our LLM-as-a-judge for all evaluations.

For embedding generation, we employed NovaSearch/stella\_en\_400M\_v5 (MIT license) from Huggingface. Our experiments were conducted across two model families with varying parameter sizes: Qwen/Qwen2.5-3B-Instruct, Qwen/Qwen2.5-14B-Instruct, and Qwen/Qwen2.5-72B-Instruct from the Qwen family, and google/gemma-3-4b-it, google/gemma-3-12b-it, and google/gemma-3-27b-it from the Gemma family.

In compliance with licensing requirements, we adhered to both the Qwen and Gemma license agreements. Qwen requires attribution by displaying "Built with Qwen" or "Improved using Qwen" when distributing AI models and special authorization for services with over 100 million monthly active users. Gemma requires adherence to their use restrictions policy and proper attribution with copies of their license agreement to recipients. Our academic research complies with these requirements, including appropriate model attribution and usage within permitted applications.

Our hardware configuration consisted of an Intel(R) Xeon(R) Gold 6448Y CPU and four NVIDIA H100 80GB HBM3 GPUs for accelerated model inference and training.

\newpage
\begin{figure}[!th]
\centering
\begin{tcolorbox}
\begin{spacing}{1.0}
\small{
You are an AI evaluator tasked with assessing the accuracy of predicted answers to questions. Your goal is to determine how well the predicted answer aligns with the expected (gold) answer and provide a numerical score.\\
\\
You will be given the following information:\\
\\
<question>\\
\{\{\$question\}\}\\
</question>\\
\\
<gold\_answer>\\
\{\{\$gold\_answer\}\}\\
</gold\_answer>\\
\\
<predicted\_answer>\\
\{\{\$predicted\_answer\}\}\\
</predicted\_answer>\\
\\
Instructions:\\
1. Carefully read the question, gold answer, and predicted answer.\\
2. Analyze the relationship between the gold answer and the predicted answer.\\
3. Consider the following criteria:\\
   - Does the predicted answer address the same topic as the gold answer?\\
   - For time-related questions, does the predicted answer refer to the same time period, even if the format differs?\\
   - Is the core information in the predicted answer consistent with the gold answer, even if expressed differently?\\
4. Assign a score from 0 to 100, where:\\
   - 0 means the predicted answer is completely unrelated or incorrect\\
   - 100 means the predicted answer perfectly matches the gold answer\\
   - Scores in between reflect partial correctness or relevance\\
5. Output your result as a single integer only. Do not use JSON or any other format.\\
\\
Important:\\
- Do not include any examples in your analysis or output.\\
- Provide only the integer score as your output, with no explanation or formatting.\\

Score:\\
}
\end{spacing}\noindent
\end{tcolorbox}
\caption{LLM-as-a-judge prompt used to evaluate response quality.}
\label{fig:llm-as-a-judge}
\end{figure}

\newpage
\onecolumn
\section{Pseudo Code}
\label{apdx:pseudo_code}

\begin{algorithm*}[!h]
    \caption{Memory-Enhanced Conversational Learning with Dynamic Clustering and Reasoning}
    \label{alg:memory_enhanced_learning}
    \begin{algorithmic}[1]
        \State \textbf{Input:} $LLM_{extract}$, $LLM_{reason}$, $LLM_{response}$, $f_{emb}$
        \State \textbf{Initialization:} $P_0 = \{\}$, $\mathcal{M} = \{\}$, $\mathcal{R} = \{\}$
        \For{$i=1, \cdots N$}
            \State \textbf{Step 1: Episodic Memory Extraction}
            \State Observe conversation session $S_i$
            \State Extract memory fragments from $S_i$: 
            $$\{m_i^1, \cdots m_i^{n_i}\} \gets LLM_{extract}(S_i)$$
            
            \State Embed memory fragments: $\{e_i^j\}_{j=1}^{n_i}$ where $e_i^j = f_{emb}(m_i^j)$
            \State Cluster fragments into $C_i = \{c_1, \dots, c_{k_i}\}$ \Comment{using silhouette scores}

            \State Construct a set $CP_i$: \Comment{using cosine similarity}
            $$CP_i = \{(p,c): sim(p,c) > \theta, p \in P_{i-1}, c \in C_i\}$$
            \State \textbf{Step 2: Pre-Storage Memory Reasoning}
            \For{$(p,c) \in CP_i$}
                \State $M_p \gets$ memory fragments in cluster $p$
                \State $M_c \gets$ memory fragments in cluster $c$.
                \State Generate reasoning 
                $$ \{r_{p,c}^{j}\}_{j=1}^{d_{p,c}} \gets LLM_{reason}(M_p, M_c)$$
                \State Store reasoning memory fragments: $ \mathcal{R} \gets \mathcal{R} \cup \{r_{p,c}^1, \cdots, r_{p,c}^{d_{p,c}} \}$    
                \State Update $P_i$:
                $$P_i = P_{i-1} \setminus \{p : \exists c  \;\; s.t. \;\; (p,c) \in CP_i\} \cup C_i$$   
            \EndFor
            \State Store raw memory fragments: $\mathcal{M} \gets \mathcal{M} \cup \{m_i^1,\dots,m_i^{n_i}\}$
            
        \EndFor

        \State \textbf{Inference Phase}
        \State Get user query $q$, compute $e_q \gets f_{\text{emb}}(q)$
        \State Retrieve top-$k$ by similarity over $\mathcal{M} \cup \mathcal{R}$:
        $$context \gets \text{TopK}_{k}\big(\mathcal{M}\cup\mathcal{R}; \mathrm{sim}(f_{\text{emb}}(\cdot), e_q)\big)$$
        \State Generate answer: $response \gets LLM_{response}(context, q)$
        \State \textbf{Output:} $response$
        
    \end{algorithmic}
\end{algorithm*}
\twocolumn

%% file: prompts/step_1.tex
\begin{figure*}[htbp]
\centering
\begin{tcolorbox}
\begin{spacing}{1.0}
\scriptsize{
\textbf{GOAL}\\
Analyze the entire provided $\langle$Conversation$\rangle$ to identify all statements revealing personal information about the user. Categorize each piece of information as Factual, Experiential, or Subjective, and output the results as a single structured JSON object according to the $\langle$Final Output JSON Format$\rangle$.\\
\\
\textbf{INFORMATION CATEGORY DEFINITIONS}\\
1.  \textbf{Factual Information:} Objective, verifiable facts about the user's state, attributes, possessions, knowledge, skills, and relationships with others (family, friends, pets, etc.). ('What I am / What I have / Who I know')\\
    * \textit{Keywords:} is, am, have, own, live in, work as, know (skill/fact/person), my name/age/job/sister/friend is, etc.\\
    * \textit{Examples:} "My name is Alex.", "I live in New York.", "I have a Bachelor's degree in CS.", "I own two bikes.", "Emily is my sister.", "Luna is my cat.", "I know Python."\\
\\
2.  \textbf{Experiential Information:} Specific events, actions, activities, or interactions experienced by the user over time, often situated in a context. ('What I did / What happened to me')\\
    * \textit{Keywords:} went, did, saw, met, visited, learned (an experience), attended, bought (as an event), have been, have visited, have tried, have experienced, last year, yesterday, when I was..., etc.\\
    * \textit{Examples:} "I travelled to LA last weekend.", "I've assembled the IKEA bookshelf.", "I've been to Japan twice.", "I have met with the CEO.", "I attended the Imagine Dragons concert."\\
\\
3.  \textbf{Subjective Information:} The user's internal states, including preferences, habits, opinions, beliefs, goals, plans, feelings, etc. ('What I like / think / want / feel / usually do')\\
    * \textit{Keywords:} like, love, hate, prefer, think, believe, feel, want, plan to, hope to, usually, often, my goal is, etc.\\
    * \textit{Examples:} "I love spicy food.", "I usually wake up at 7 AM.", "I thought that movie was great.", "My goal is to learn Spanish.", "I want to visit Europe next year."\\
\\
\textbf{INSTRUCTIONS}\\
1. Carefully read and analyze the entire $\langle$Conversation$\rangle$. $\langle$Conversation$\rangle$ consists of messages, each containing a [message\_id] followed by its content.\\
2. Identify all specific pieces of information about the user that fall into the Factual, Experiential, or Subjective categories based on the definitions above.\\
3. Format each value as a phrase that starts with a verb in present tense, regardless of the original tense in the conversation.\\
4. For the "date" field:\\
    * For ongoing facts or current states, use the date of the message\\
    * For past events with a specific timeframe mentioned (e.g., "yesterday", "three days ago"), calculate and use the actual date based on the message date\\
    * For past events mentioned in the conversation, mark as "Before [message-date]"\\
    * For future plans or intentions, mark as "After [message-date]"\\
5. Format the output as a single JSON object with three categories: "Factual\_Information", "Experiential\_Information", and "Subjective\_Information". Use empty lists ([]) for categories with no information.\\
6. Use the exact same [message\_id] as in the original message. Do not include pronouns in the value.\\
\\
\textbf{Example}\\
$\langle$Conversation$\rangle$\\

[msg-301] (2024-05-17 Friday) I'm living in Rome now with my girlfriend, Hana. We moved here last summer because she started grad school at Stanford.

[msg-302] (2024-05-17 Friday) I quit my job at Coupang in March. I just didn't see myself growing there anymore.

[msg-303] (2024-05-17 Friday) I'm thinking about switching into UX design. I've always liked the idea of making tech more human-friendly.

[msg-304] (2024-05-17 Friday) My brother Junho lives in Seattle. He's an engineer and always sends me photos of the mountains.

[msg-305] (2024-05-17 Friday) I ate chicken with my friends yesterday.

Answer:
\begin{verbatim}
{
  "Factual\_Information": [
    {
      "key": "current residence",
      "value": "Lives in Rome with girlfriend Hana",
      "message_id": "msg-301",
      "date": "2024-05-17"
    },...
  ],
  "Experiential\_Information": [
    {
      "key": "job resignation",
      "value": "Quit job at Coupang in March",
      "message_id": "msg-302",
      "date": "Before 2024-05-17"
    },...
  ],
  "Subjective_Information": [
    {
      "key": "career dissatisfaction",
      "value": "Be Felt no growth potential at Coupang",
      "message\_id": "msg-302",
      "date": "Before 2024-05-17"
    },...
  ]
}\end{verbatim}

$\langle$Conversation$\rangle$\\
\{\{\$conversation\}\}\\
}
\end{spacing}\noindent
\end{tcolorbox}
\caption{Step 1: Personal information extraction and categorization prompt.}
\label{fig:step1-prompt}
\end{figure*}

%% file: prompts/step_2.tex
\begin{figure*}[htbp]
\centering
\begin{tcolorbox}
\begin{spacing}{1.0}
\scriptsize{
You are an AI assistant analyzing memory fragments to generate insights. Your task is to identify patterns and connections from the data provided.\\
\\
Analyze these fragments and generate insights based on five inference types:\\
- Extension/Generalization: The process of expanding information from specific cases or situations to broader categories, domains, or patterns. This type of inference derives more general characteristics or tendencies from concrete information.\\
- Accumulation: The process of identifying behaviors, experiences, or patterns that repeat or persist over time. This type of inference focuses on frequency, consistency, and persistence to infer habitual patterns or significant trends.\\
- Specification/Refinement: The process of breaking down general information into more detailed and specific aspects. This type of inference decomposes broad concepts or experiences into concrete elements or details.\\
- Transformation: The process of identifying changes in states, perspectives, emotions, behaviors, etc. over time. This type of inference discerns transitions or developments between previous and current/new states.\\
- Connection/Implication: The process of identifying relationships, causality, or meaning between seemingly disparate pieces of information. This type of inference discerns connections or conclusions not explicitly mentioned.\\
\\
Your output should be formatted as a JSON object with an "extended\_insight" key containing an array of inference objects. Each inference object should have the following structure:\\
\begin{verbatim}
{
  "inference_type": "one of the five inference types",
  "key": "brief description of the insight",
  "date": "relevant date or date range",
  "value": "detailed description of the insight (12 words or less)"
}
\end{verbatim}

Important instructions:\\
- You do NOT need to use all five inference types. Select only the inference types that clearly apply to the data.\\
- Include multiple different inference types when appropriate, but don't force all five types.\\
- You may use the same inference type multiple times for different insights if appropriate.\\
- Focus on quality over quantity - provide meaningful insights based on the data.\\
- Avoid trivial or insignificant insights - focus only on substantive patterns and connections.\\
\\
$\langle$example$\rangle$\\
Example:\\
Below are the memory fragments to analyze:

[tech purchase, 2023-03-05]: Jordan buy new drawing tablet

[software usage, 2023-03-07]: Jordan spend three hours learning Procreate app

[online activity, 2023-03-15]: Jordan create account on digital art community DeviantArt

[social media, 2023-03-22]: Jordan share first digital artwork on Instagram
\\
Output:\\
\begin{verbatim}
{
  "extended_insight": [
    {
      "inference_type": "extension/generalization",
      "key": "skill development approach",
      "date": "2023-03-05 to 2023-03-22",
      "value": "Jordan follows a methodical learning approach with appropriate tools"
    },
    {
      "inference_type": "accumulation",
      "key": "digital art activities",
      "date": "2023-03-05 to 2023-03-22",
      "value": "Jordan engaged in 4 digital art activities within 17 days"
    },
    {
      "inference_type": "specification/refinement",
      "key": "artistic medium",
      "date": "2023-03-22",
      "value": "Jordan uses tablet-based digital illustration with Procreate"
    },
    {
      "inference_type": "transformation",
      "key": "identity shift",
      "date": "Before 2023-03-05 to 2023-03-22",
      "value": "Jordan evolved from art appreciator to digital artist"
    },
    {
      "inference_type": "connection/implication",
      "key": "artistic background",
      "date": "Before 2023-03-05",
      "value": "Jordan likely has previous art experience"
    }
  ]
}
\end{verbatim}
$\langle$/example$\rangle$\\
\\
Below are the memory fragments to analyze:\\
\{\{\$memory\_fragments\}\}
}
\end{spacing}\noindent
\end{tcolorbox}
\caption{Step 2: Memory pattern analysis and inference prompt.}
\label{fig:step2-prompt}
\end{figure*}

%% file: table/all_table.tex
\begingroup
\begin{table}[!hb]
\centering
\resizebox{\textwidth}{!}{%
\renewcommand{\arraystretch}{0.9}     
\fontsize{6}{7}\selectfont
\setlength\tabcolsep{3pt}
\begin{tabular}{
  cc   
  c    
  >{\columncolor{gray!20}}c	
  c	
  c	
  c	
  c	
  c	
  c	
  >{\columncolor{gray!20}}c	
  c	
  c	
  c	
  c	
  c	
  c	
}
\toprule
\multicolumn{2}{c}{\multirow{3}{*}{\makecell[c]{\textbf{Inference}\\\textbf{LLM}}}}
  & \multirow{3}{*}{\makecell[c]{\textbf{Model}}}
  & \multicolumn{7}{c}{\textbf{LongMemEval}}
  & \multicolumn{7}{c}{\textbf{LoCoMo}}\\
\cmidrule(lr){4-10}\cmidrule(lr){11-17}
 &  & &\multicolumn{1}{c}{\textbf{LLM}} & ROUGE-1 & ROUGE-L & BLEU-1 & METEOR & BERTScore & token length
 &\multicolumn{1}{c}{\textbf{LLM}} & ROUGE-1 & ROUGE-L & BLEU-1 & METEOR & BERTScore & token length \\
\midrule

\multirow{33}{*}{\rotatebox{90}{Qwen2.5}}
& \multirow{11}{*}{3B}
& Zero & 15.93 & 7.77 & 7.20 & 5.16 & 4.30 & 85.44 & 0.00 & 22.24 & 7.12 & 6.10 & 6.12 & 4.85 & 83.99 & 0.00 \\
 &  & Full & 32.55 & 19.25 & 18.97 & 16.83 & 11.60 & 89.34 & 18710.50 & 42.56 & 20.33 & 19.50 & 15.87 & 16.37 & 86.21 & 19643.80 \\
 &  & Turn & 38.43 & 23.71 & 23.23 & 20.24 & 13.80 & 89.58 & 1854.30 & 44.90 & 22.06 & 21.17 & 16.31 & 16.57 & 86.49 & 2009.30 \\
 &  & Session & 31.02 & 19.38 & 18.88 & 16.10 & 10.76 & 88.42 & 1919.80 & 41.40 & 20.77 & 19.96 & 15.75 & 15.44 & 86.15 & 1989.30 \\
 &  & Segment & 36.59 & 21.86 & 21.32 & 19.53 & 12.90 & 89.35 & 1770.79 & 46.28 & 25.29 & 24.22 & 18.94 & 19.44 & 86.43 & 1881.45 \\
 &  & SeCom & 34.52 & 23.15 & 22.70 & 19.97 & 13.51 & 89.06 & 1775.10 & 42.93 & 23.85 & 22.85 & 17.75 & 17.60 & 86.23 & 1884.30 \\
 &  & HippoRAG-2 & 40.45 & 27.31 & 26.72 & 24.26 & 16.27 & 89.46 & 3811.61 & 46.74 & 26.82 & 25.93 & 21.68 & 19.99 & 87.26 & 3811.61 \\
 &  & A-Mem & 44.42 & 31.15 & 30.27 & 27.64 & 20.31 & 90.06 & 6199.24 & 42.07 & 29.02 & 28.44 & 23.70 & 21.85 & 88.31 & 6199.24 \\
 &  & PREMem & 50.80 & 33.81 & 33.45 & 30.49 & 21.76 & 90.84 & 2032.40 & 53.79 & 24.64 & 23.16 & 15.86 & 16.46 & 86.07 & 2033.90 \\
 &  & \textit{with bm25} & 48.25 & 33.26 & 32.82 & 29.78 & 22.00 & 90.64 & 2032.20 & 51.20 & 22.69 & 21.18 & 14.13 & 14.71 & 85.85 & 2034.50 \\
 &  & PREMem\_small & 46.13 & 29.14 & 28.51 & 25.92 & 18.83 & 90.20 & 2032.30 & 51.26 & 23.15 & 21.55 & 14.49 & 15.01 & 85.96 & 2034.00 \\
\cmidrule(lr){2-17}

& \multirow{11}{*}{14B}
& Zero & 16.08 & 10.10 & 9.90 & 7.49 & 5.55 & 85.05 & 0.00 & 25.19 & 7.12 & 5.45 & 6.10 & 6.22 & 82.84 & 0.00 \\
 &  & Full & 37.43 & 25.37 & 24.73 & 19.92 & 15.84 & 88.13 & 18710.50 & 60.57 & 24.08 & 22.81 & 18.16 & 21.85 & 86.62 & 19643.80 \\
 &  & Turn & 39.70 & 25.27 & 24.51 & 20.48 & 15.68 & 87.88 & 1854.30 & 61.63 & 28.90 & 27.49 & 22.87 & 22.34 & 87.77 & 2009.30 \\
 &  & Session & 29.04 & 19.27 & 18.79 & 14.82 & 11.41 & 86.71 & 1919.80 & 54.46 & 25.68 & 24.34 & 20.27 & 20.49 & 86.48 & 1989.30 \\
 &  & Segment & 39.02 & 24.62 & 24.02 & 20.15 & 14.92 & 87.80 & 1770.79 & 63.54 & 33.22 & 31.90 & 26.16 & 26.12 & 87.66 & 1881.45 \\
 &  & SeCom & 37.63 & 24.51 & 23.90 & 19.33 & 14.56 & 87.87 & 1775.10 & 60.42 & 31.04 & 29.64 & 24.59 & 23.95 & 87.32 & 1884.30 \\
 &  & HippoRAG-2 & 44.69 & 29.24 & 28.44 & 22.73 & 19.51 & 87.84 & 3811.61 & 61.67 & 30.41 & 29.02 & 24.75 & 24.19 & 87.38 & 3811.61 \\
 &  & A-Mem & 50.32 & 32.99 & 31.92 & 27.17 & 23.41 & 88.54 & 6199.24 & 43.62 & 30.24 & 29.66 & 25.38 & 22.51 & 88.27 & 6199.24 \\
 &  & PREMem & 64.73 & 40.41 & 39.86 & 32.16 & 28.75 & 89.54 & 2032.40 & 68.03 & 29.42 & 27.14 & 21.50 & 21.59 & 86.87 & 2033.90 \\
 &  & \textit{with bm25} & 59.37 & 38.66 & 38.19 & 30.20 & 26.88 & 89.22 & 2032.20 & 63.97 & 26.67 & 24.43 & 18.89 & 19.58 & 86.41 & 2035.00 \\
 &  & PREMem\_small & 51.86 & 31.06 & 30.64 & 24.60 & 21.33 & 88.42 & 2032.30 & 64.52 & 27.21 & 25.03 & 19.54 & 19.12 & 86.57 & 2034.00 \\
 \cmidrule(lr){2-17}

& \multirow{11}{*}{72B}
& Zero & 20.61 & 12.59 & 12.27 & 10.32 & 6.83 & 86.29 & 0.00 & 25.15 & 8.43 & 6.72 & 6.94 & 8.01 & 83.07 & 0.00 \\
 &  & Full & 36.95 & 24.53 & 23.94 & 20.20 & 14.93 & 88.12 & 18710.50 & 63.97 & 20.83 & 19.50 & 14.92 & 20.77 & 85.94 & 19643.80 \\
 &  & Turn & 40.63 & 26.22 & 25.57 & 20.98 & 15.30 & 88.33 & 1854.30 & 63.71 & 26.28 & 24.85 & 20.26 & 21.20 & 87.27 & 2009.30 \\
 &  & Session & 30.89 & 20.76 & 20.31 & 16.54 & 11.90 & 87.38 & 1919.80 & 54.21 & 23.85 & 22.64 & 18.75 & 19.47 & 86.27 & 1989.30 \\
 &  & Segment & 38.61 & 25.60 & 25.13 & 21.11 & 15.09 & 88.48 & 1770.79 & 61.91 & 29.72 & 28.30 & 23.40 & 23.58 & 87.20 & 1881.45 \\
 &  & SeCom & 39.40 & 25.25 & 24.82 & 20.78 & 14.59 & 88.24 & 1775.10 & 58.51 & 28.02 & 26.54 & 22.06 & 22.22 & 86.90 & 1884.30 \\
 &  & HippoRAG-2 & 45.95 & 29.79 & 29.27 & 23.35 & 18.84 & 88.04 & 3811.61 & 61.62 & 30.40 & 29.07 & 24.43 & 24.12 & 87.43 & 3811.61 \\
 &  & A-Mem & 53.58 & 36.25 & 35.42 & 29.46 & 25.43 & 89.34 & 6199.24 & 45.59 & 31.68 & 30.94 & 26.35 & 23.08 & 88.56 & 6199.24 \\
 &  & PREMem & 67.50 & 45.36 & 44.89 & 36.12 & 30.90 & 90.19 & 2032.40 & 70.96 & 27.05 & 24.71 & 19.05 & 20.48 & 86.47 & 2033.90 \\
 &  & \textit{with bm25} & 62.03 & 42.10 & 41.67 & 34.21 & 28.76 & 89.98 & 2032.20 & 66.96 & 25.05 & 22.85 & 17.56 & 18.52 & 86.23 & 2034.50 \\
 &  & PREMem\_small & 56.89 & 35.44 & 34.97 & 27.95 & 24.07 & 89.13 & 2032.30 & 66.68 & 25.59 & 23.39 & 17.87 & 18.73 & 86.39 & 2034.00 \\

\midrule

\multirow{33}{*}{\rotatebox{90}{gemma-3}}
& \multirow{11}{*}{4B}
& Zero & 18.32 & 11.35 & 10.99 & 10.20 & 5.50 & 86.25 & 1.00 & 20.91 & 11.33 & 10.89 & 7.54 & 4.29 & 86.43 & 1.00 \\
 &  & Full & 32.45 & 20.81 & 20.48 & 17.15 & 11.93 & 88.14 & 18669.20 & 42.49 & 26.32 & 25.26 & 19.46 & 15.77 & 87.58 & 19591.00 \\
 &  & Turn & 34.96 & 22.19 & 21.80 & 17.37 & 13.01 & 88.42 & 1838.90 & 46.53 & 26.78 & 25.96 & 19.05 & 14.94 & 87.92 & 1984.00 \\
 &  & Session & 28.31 & 18.28 & 17.94 & 14.37 & 9.59 & 87.90 & 1929.00 & 38.22 & 23.54 & 22.95 & 16.90 & 13.53 & 87.09 & 1967.40 \\
 &  & Segment & 36.49 & 21.00 & 20.65 & 16.90 & 12.74 & 88.23 & 1843.93 & 47.79 & 31.16 & 30.28 & 21.55 & 17.67 & 87.79 & 1969.70 \\
 &  & SeCom & 35.24 & 21.16 & 20.86 & 16.86 & 12.78 & 88.06 & 1844.72 & 45.15 & 29.68 & 28.91 & 21.66 & 16.58 & 87.72 & 1971.58 \\
 &  & HippoRAG-2 & 41.00 & 24.47 & 24.04 & 19.22 & 15.68 & 88.71 & 4000.39 & 47.48 & 31.14 & 30.21 & 23.81 & 18.86 & 88.27 & 4000.39 \\
 &  & A-Mem & 42.30 & 28.86 & 28.49 & 22.23 & 19.85 & 88.42 & 6250.81 & 45.70 & 33.11 & 32.66 & 28.11 & 25.70 & 88.84 & 6250.81 \\
 &  & PREMem & 53.39 & 31.63 & 31.37 & 25.00 & 23.84 & 89.03 & 1962.70 & 50.14 & 29.00 & 27.29 & 19.60 & 17.01 & 86.98 & 1957.80 \\
 &  & \textit{with bm25} & 49.71 & 30.81 & 30.68 & 25.13 & 22.76 & 89.09 & 1964.60 & 48.40 & 26.41 & 24.78 & 17.32 & 14.65 & 86.86 & 1961.50 \\
 &  & PREMem\_small & 48.50 & 30.08 & 29.65 & 24.94 & 21.70 & 89.36 & 1963.70 & 47.83 & 27.24 & 25.60 & 18.10 & 16.11 & 86.78 & 1956.30 \\
\cmidrule(lr){2-17}

& \multirow{11}{*}{12B}
& Zero & 27.72 & 18.05 & 17.91 & 14.22 & 7.38 & 88.57 & 1.00 & 21.28 & 11.19 & 10.84 & 7.36 & 4.26 & 85.66 & 1.00 \\
 &  & Full & 34.80 & 21.40 & 21.25 & 17.25 & 12.50 & 88.23 & 18669.20 & 43.68 & 27.62 & 27.13 & 21.06 & 18.73 & 88.00 & 19591.00 \\
 &  & Turn & 36.60 & 22.45 & 22.21 & 17.50 & 13.15 & 88.09 & 1838.90 & 45.46 & 27.05 & 26.54 & 20.59 & 18.89 & 87.91 & 1984.00 \\
 &  & Session & 28.76 & 16.93 & 16.68 & 13.23 & 9.64 & 87.74 & 1929.00 & 37.95 & 24.19 & 23.80 & 17.85 & 16.16 & 86.96 & 1967.40 \\
 &  & Segment & 34.15 & 21.09 & 20.81 & 17.35 & 12.38 & 87.92 & 1843.93 & 46.72 & 31.31 & 30.79 & 23.73 & 22.12 & 87.92 & 1969.70 \\
 &  & SeCom & 37.35 & 23.52 & 23.37 & 19.44 & 13.83 & 88.31 & 1844.72 & 44.80 & 30.09 & 29.67 & 22.82 & 20.17 & 87.61 & 1971.58 \\
 &  & HippoRAG-2 & 43.90 & 27.14 & 26.77 & 21.35 & 17.74 & 88.71 & 4000.39 & 44.28 & 29.66 & 29.17 & 22.51 & 20.07 & 87.79 & 4000.39 \\
 &  & A-Mem & 38.98 & 28.06 & 27.63 & 22.45 & 19.73 & 87.88 & 6250.81 & 43.94 & 31.18 & 30.81 & 25.95 & 23.63 & 88.47 & 6250.81 \\
 &  & PREMem & 57.70 & 34.44 & 33.93 & 27.73 & 25.58 & 89.20 & 1962.70 & 49.99 & 30.07 & 28.85 & 21.83 & 19.21 & 87.54 & 1957.80 \\
 &  & \textit{with bm25} & 53.70 & 32.08 & 31.76 & 26.47 & 24.60 & 88.97 & 1964.60 & 48.16 & 26.67 & 25.62 & 19.12 & 16.98 & 87.17 & 1961.50 \\
 &  & PREMem\_small & 55.01 & 31.56 & 31.14 & 25.46 & 22.72 & 88.74 & 1963.70 & 47.56 & 27.90 & 26.79 & 19.62 & 17.83 & 87.24 & 1956.30 \\
 \cmidrule(lr){2-17}

& \multirow{11}{*}{27B}
& Zero & 30.15 & 17.58 & 17.43 & 12.88 & 5.95 & 88.62 & 1.00 & 22.16 & 10.19 & 9.85 & 6.68 & 5.20 & 85.63 & 1.00 \\
 &  & Full & 35.90 & 22.55 & 22.18 & 18.51 & 12.11 & 88.72 & 18669.20 & 47.37 & 29.36 & 28.71 & 22.17 & 18.38 & 88.16 & 19591.00 \\
 &  & Turn & 37.98 & 23.62 & 23.25 & 18.69 & 13.02 & 88.69 & 1838.90 & 49.72 & 27.53 & 26.76 & 21.01 & 17.87 & 87.99 & 1984.00 \\
 &  & Session & 27.58 & 16.76 & 16.50 & 13.07 & 7.77 & 88.31 & 1929.00 & 43.32 & 24.57 & 23.87 & 18.32 & 15.95 & 86.97 & 1967.40 \\
 &  & Segment & 36.83 & 21.54 & 21.11 & 16.92 & 10.17 & 88.64 & 1843.93 & 50.49 & 31.47 & 30.74 & 23.97 & 20.94 & 87.96 & 1969.70 \\
 &  & SeCom & 38.93 & 23.16 & 22.91 & 18.22 & 11.28 & 88.65 & 1844.72 & 49.11 & 30.21 & 29.51 & 22.95 & 18.68 & 87.77 & 1971.58 \\
 &  & HippoRAG-2 & 43.15 & 27.34 & 27.00 & 20.42 & 14.02 & 89.46 & 4000.39 & 49.49 & 30.59 & 29.79 & 23.25 & 19.35 & 87.89 & 4000.39 \\
 &  & A-Mem & 45.32 & 31.90 & 31.33 & 26.27 & 21.79 & 88.60 & 6250.81 & 44.47 & 32.76 & 32.23 & 27.08 & 23.75 & 88.73 & 6250.81 \\
 &  & PREMem & 61.86 & 39.20 & 38.81 & 30.85 & 25.50 & 89.88 & 1962.70 & 54.55 & 30.61 & 28.97 & 22.16 & 19.89 & 87.42 & 1957.80 \\
 &  & \textit{with bm25} & 56.01 & 36.70 & 36.46 & 29.54 & 24.42 & 89.85 & 1964.60 & 52.89 & 27.74 & 26.11 & 19.54 & 17.38 & 87.18 & 1961.50 \\
 &  & PREMem\_small & 57.15 & 36.11 & 35.44 & 28.73 & 23.08 & 89.77 & 1963.70 & 53.13 & 29.24 & 27.57 & 21.01 & 19.03 & 87.22 & 1956.30 \\

\midrule

\multirow{33}{*}{\rotatebox{90}{gpt-4.1}}
& \multirow{11}{*}{nano}
& Zero & 11.36 & 6.82 & 6.68 & 5.24 & 3.52 & 84.51 & 0.00 & 24.06 & 10.89 & 10.51 & 7.61 & 6.30 & 85.39 & 0.00 \\
 &  & Full & 38.27 & 24.60 & 24.13 & 21.20 & 15.47 & 89.19 & 18691.50 & 47.50 & 29.40 & 28.49 & 24.02 & 22.15 & 88.11 & 19651.60 \\
 &  & Turn & 37.12 & 24.45 & 24.11 & 20.04 & 14.68 & 89.61 & 1858.00 & 50.33 & 30.37 & 29.59 & 24.23 & 21.96 & 88.47 & 2013.10 \\
 &  & Session & 27.28 & 17.24 & 16.95 & 13.65 & 10.08 & 88.54 & 1912.30 & 44.93 & 27.57 & 26.99 & 22.06 & 19.92 & 87.40 & 2000.20 \\
 &  & Segment & 37.43 & 23.90 & 23.42 & 19.97 & 14.59 & 89.63 & 1644.89 & 52.15 & 34.67 & 33.75 & 27.61 & 25.22 & 88.43 & 1701.81 \\
 &  & SeCom & 37.44 & 24.98 & 24.63 & 20.64 & 15.45 & 89.53 & 1647.14 & 48.10 & 31.06 & 30.39 & 24.93 & 21.96 & 87.91 & 1708.30 \\
 &  & HippoRAG-2 & 43.05 & 29.02 & 28.71 & 24.45 & 18.68 & 89.67 & 3667.23 & 50.40 & 32.82 & 32.13 & 26.80 & 24.26 & 88.09 & 3667.23 \\
 &  & A-Mem & 51.83 & 34.73 & 33.68 & 28.69 & 23.04 & 89.14 & 5902.39 & 50.45 & 35.91 & 35.13 & 30.14 & 26.16 & 89.13 & 5902.39 \\
 &  & PREMem & 58.75 & 39.66 & 39.11 & 32.77 & 25.17 & 90.87 & 2033.70 & 58.75 & 32.93 & 31.41 & 24.04 & 20.97 & 87.59 & 2035.80 \\
 &  & \textit{with bm25} & 57.34 & 39.51 & 39.31 & 33.28 & 26.04 & 90.73 & 2033.80 & 56.76 & 30.96 & 29.61 & 22.32 & 19.42 & 87.31 & 2028.80 \\
 &  & PREMem\_small & 55.64 & 38.41 & 38.03 & 32.26 & 25.61 & 91.00 & 2034.10 & 54.81 & 32.06 & 30.78 & 23.00 & 20.40 & 87.51 & 2035.90 \\
\cmidrule(lr){2-17}

& \multirow{11}{*}{mini}
& Zero & 9.88 & 5.40 & 5.18 & 3.45 & 2.88 & 84.01 & 0.00 & 23.56 & 10.99 & 10.40 & 8.22 & 6.54 & 84.71 & 0.00 \\
 &  & Full & 44.13 & 27.48 & 27.12 & 22.69 & 17.88 & 88.53 & 18691.50 & 56.41 & 33.68 & 32.72 & 27.16 & 26.54 & 88.30 & 19651.60 \\
 &  & Turn & 39.51 & 25.38 & 24.87 & 20.13 & 16.16 & 88.07 & 1858.00 & 54.71 & 30.33 & 29.47 & 23.97 & 23.44 & 88.18 & 2013.10 \\
 &  & Session & 29.71 & 18.01 & 17.88 & 13.63 & 10.77 & 87.21 & 1912.30 & 48.14 & 27.53 & 26.87 & 21.99 & 20.48 & 87.27 & 2000.20 \\
 &  & Segment & 41.24 & 25.86 & 25.35 & 20.29 & 16.85 & 88.21 & 1644.89 & 53.29 & 33.17 & 32.28 & 26.15 & 24.27 & 87.76 & 1701.81 \\
 &  & SeCom & 42.32 & 26.77 & 26.37 & 21.69 & 17.02 & 88.54 & 1647.14 & 53.35 & 33.17 & 32.28 & 26.15 & 24.27 & 87.76 & 1708.30 \\
 &  & HippoRAG-2 & 44.78 & 29.08 & 28.67 & 22.61 & 19.51 & 88.17 & 3667.23 & 54.63 & 34.11 & 33.16 & 28.17 & 26.15 & 88.27 & 3667.23 \\
 &  & A-Mem & 53.94 & 35.48 & 34.43 & 28.61 & 24.77 & 88.88 & 5902.39 & 52.73 & 37.05 & 36.18 & 31.60 & 28.83 & 89.32 & 5902.39 \\
 &  & PREMem & 67.61 & 43.19 & 42.43 & 33.88 & 30.55 & 89.64 & 2033.70 & 64.88 & 34.50 & 32.57 & 25.41 & 22.98 & 87.59 & 2035.80 \\
 &  & \textit{with bm25} & 65.63 & 41.23 & 40.70 & 32.34 & 29.47 & 89.32 & 2033.80 & 62.40 & 31.94 & 30.10 & 23.14 & 20.40 & 87.26 & 2036.50 \\
 &  & PREMem\_small & 66.80 & 42.44 & 41.89 & 33.86 & 30.33 & 89.75 & 2034.10 & 58.98 & 32.50 & 30.81 & 23.66 & 22.11 & 87.45 & 2035.90 \\
\cmidrule(lr){2-17}

& \multirow{11}{*}{base}
& Zero & 9.01 & 4.20 & 4.11 & 2.27 & 2.37 & 83.24 & 0.00 & 22.76 & 10.62 & 9.74 & 8.59 & 6.26 & 84.73 & 0.00 \\
 &  & Full & 39.57 & 24.09 & 23.73 & 18.73 & 14.23 & 87.77 & 18691.50 & 58.98 & 33.95 & 32.64 & 27.40 & 26.97 & 88.36 & 19651.60 \\
 &  & Turn & 40.69 & 25.22 & 24.87 & 20.87 & 16.18 & 87.93 & 1858.00 & 57.09 & 31.34 & 30.11 & 25.22 & 24.14 & 88.40 & 2013.10 \\
 &  & Session & 30.26 & 18.29 & 18.03 & 14.32 & 10.48 & 86.69 & 1912.30 & 50.12 & 27.91 & 27.07 & 22.53 & 21.09 & 87.26 & 2000.20 \\
 &  & Segment & 41.18 & 24.92 & 24.42 & 19.66 & 15.28 & 87.99 & 1644.89 & 58.67 & 35.83 & 34.75 & 29.15 & 27.79 & 88.29 & 1701.81 \\
 &  & SeCom & 41.98 & 26.20 & 25.81 & 21.14 & 16.34 & 88.05 & 1647.14 & 56.67 & 34.96 & 33.79 & 28.53 & 26.19 & 88.13 & 1708.30 \\
 &  & HippoRAG-2 & 45.19 & 29.19 & 28.79 & 22.44 & 19.20 & 87.96 & 3667.23 & 57.31 & 34.04 & 32.91 & 28.47 & 26.59 & 88.26 & 3667.23 \\
 &  & A-Mem & 55.91 & 37.50 & 36.41 & 30.94 & 26.07 & 89.48 & 5902.39 & 49.52 & 34.73 & 33.87 & 29.21 & 25.88 & 88.96 & 5902.39 \\
 &  & PREMem & 71.38 & 44.57 & 44.07 & 35.67 & 32.01 & 89.74 & 2033.70 & 67.73 & 35.92 & 33.76 & 27.35 & 24.31 & 87.88 & 2035.80 \\
 &  & \textit{with bm25} & 67.44 & 42.43 & 42.06 & 34.23 & 31.06 & 89.39 & 2033.80 & 64.99 & 32.93 & 30.75 & 24.60 & 21.61 & 87.52 & 2036.50 \\
 &  & PREMem\_small & 68.08 & 43.24 & 42.64 & 34.10 & 30.47 & 89.80 & 2034.10 & 60.74 & 32.84 & 31.06 & 24.85 & 22.48 & 87.68 & 2035.90 \\
\bottomrule
\end{tabular}
}
\caption{Complete experimental results.}
\label{tab:all}
\end{table}
\endgroup

%% file: acl_latex.bbl
\begin{thebibliography}{54}
\providecommand{\natexlab}[1]{#1}

\bibitem[{Anderson(2013)}]{anderson2013architecture}
John~R Anderson. 2013.
\newblock \emph{The architecture of cognition}.
\newblock Psychology Press.

\bibitem[{Bae et~al.(2022)Bae, Kwak, Kang, Lee, Kim, Jeong, Kim, Lee, Park, and Sung}]{bae-etal-2022-keep-me-update}
Sanghwan Bae, Donghyun Kwak, Soyoung Kang, Min~Young Lee, Sungdong Kim, Yuin Jeong, Hyeri Kim, Sang-Woo Lee, Woomyoung Park, and Nako Sung. 2022.
\newblock \href {https://doi.org/10.18653/v1/2022.findings-emnlp.276} {Keep me updated! memory management in long-term conversations}.
\newblock In \emph{Findings of the Association for Computational Linguistics: EMNLP 2022}, pages 3769--3787, Abu Dhabi, United Arab Emirates. Association for Computational Linguistics.

\bibitem[{Bartlett(1995)}]{bartlett1995remembering}
Frederic~Charles Bartlett. 1995.
\newblock \emph{Remembering: A study in experimental and social psychology}.
\newblock Cambridge university press.

\bibitem[{Bransford and Johnson(1972)}]{bransford1972contextual}
John~D Bransford and Marcia~K Johnson. 1972.
\newblock Contextual prerequisites for understanding: Some investigations of comprehension and recall.
\newblock \emph{Journal of verbal learning and verbal behavior}, 11(6):717--726.

\bibitem[{Carey(1985)}]{carey1985conceptual}
Susan Carey. 1985.
\newblock Conceptual change in childhood.
\newblock \emph{(No Title)}.

\bibitem[{Chen et~al.(2025)Chen, Li, Chang, Huang, Wang, and Li}]{chen-etal-2025-compress}
Nuo Chen, Hongguang Li, Jianhui Chang, Juhua Huang, Baoyuan Wang, and Jia Li. 2025.
\newblock \href {https://aclanthology.org/2025.coling-main.51/} {Compress to impress: Unleashing the potential of compressive memory in real-world long-term conversations}.
\newblock In \emph{Proceedings of the 31st International Conference on Computational Linguistics}, pages 755--773, Abu Dhabi, UAE. Association for Computational Linguistics.

\bibitem[{Chi(2009)}]{chi2009three}
Michelene~TH Chi. 2009.
\newblock Three types of conceptual change: Belief revision, mental model transformation, and categorical shift.
\newblock In \emph{International handbook of research on conceptual change}, pages 89--110. Routledge.

\bibitem[{Chi et~al.(1981)}]{chi1981expertise}
Michelene~TH Chi and 1 others. 1981.
\newblock Expertise in problem solving.

\bibitem[{Chu et~al.(2024)Chu, Chen, Chen, Yu, Wang, Liu, and Qin}]{chu-etal-2024-timebench}
Zheng Chu, Jingchang Chen, Qianglong Chen, Weijiang Yu, Haotian Wang, Ming Liu, and Bing Qin. 2024.
\newblock \href {https://doi.org/10.18653/v1/2024.acl-long.66} {{T}ime{B}ench: A comprehensive evaluation of temporal reasoning abilities in large language models}.
\newblock In \emph{Proceedings of the 62nd Annual Meeting of the Association for Computational Linguistics (Volume 1: Long Papers)}, pages 1204--1228, Bangkok, Thailand. Association for Computational Linguistics.

\bibitem[{Du et~al.(2025)Du, Huang, Zheng, Wang, Montella, Lapata, Wong, and Pan}]{du2025rethinkingmemoryaitaxonomy}
Yiming Du, Wenyu Huang, Danna Zheng, Zhaowei Wang, Sebastien Montella, Mirella Lapata, Kam-Fai Wong, and Jeff~Z. Pan. 2025.
\newblock \href {https://arxiv.org/abs/2505.00675} {Rethinking memory in ai: Taxonomy, operations, topics, and future directions}.
\newblock \emph{Preprint}, arXiv:2505.00675.

\bibitem[{Edge et~al.(2025)Edge, Trinh, Cheng, Bradley, Chao, Mody, Truitt, Metropolitansky, Ness, and Larson}]{edge2025localglobalgraphrag}
Darren Edge, Ha~Trinh, Newman Cheng, Joshua Bradley, Alex Chao, Apurva Mody, Steven Truitt, Dasha Metropolitansky, Robert~Osazuwa Ness, and Jonathan Larson. 2025.
\newblock \href {https://arxiv.org/abs/2404.16130} {From local to global: A graph rag approach to query-focused summarization}.
\newblock \emph{Preprint}, arXiv:2404.16130.

\bibitem[{Fauconnier and Turner(2008)}]{fauconnier2008way}
Gilles Fauconnier and Mark Turner. 2008.
\newblock \emph{The way we think: Conceptual blending and the mind's hidden complexities}.
\newblock Basic books.

\bibitem[{Fountas et~al.(2025)Fountas, Benfeghoul, Oomerjee, Christopoulou, Lampouras, Ammar, and Wang}]{fountas2025humaninspired}
Zafeirios Fountas, Martin Benfeghoul, Adnan Oomerjee, Fenia Christopoulou, Gerasimos Lampouras, Haitham~Bou Ammar, and Jun Wang. 2025.
\newblock \href {https://openreview.net/forum?id=BI2int5SAC} {Human-inspired episodic memory for infinite context {LLM}s}.
\newblock In \emph{The Thirteenth International Conference on Learning Representations}.

\bibitem[{Ge et~al.(2025)Ge, Romeo, Cai, Shu, Sunkara, Benajiba, and Zhang}]{ge2025tremuneurosymbolictemporalreasoning}
Yubin Ge, Salvatore Romeo, Jason Cai, Raphael Shu, Monica Sunkara, Yassine Benajiba, and Yi~Zhang. 2025.
\newblock \href {https://arxiv.org/abs/2502.01630} {Tremu: Towards neuro-symbolic temporal reasoning for llm-agents with memory in multi-session dialogues}.
\newblock \emph{Preprint}, arXiv:2502.01630.

\bibitem[{Guo et~al.(2025)Guo, Xia, Yu, Ao, and Huang}]{guo2025lightragsimplefastretrievalaugmented}
Zirui Guo, Lianghao Xia, Yanhua Yu, Tu~Ao, and Chao Huang. 2025.
\newblock \href {https://arxiv.org/abs/2410.05779} {Lightrag: Simple and fast retrieval-augmented generation}.
\newblock \emph{Preprint}, arXiv:2410.05779.

\bibitem[{Gutiérrez et~al.(2024)Gutiérrez, Shu, Gu, Yasunaga, and Su}]{gutierrez2024hipporag}
Bernal~Jiménez Gutiérrez, Yiheng Shu, Yu~Gu, Michihiro Yasunaga, and Yu~Su. 2024.
\newblock \href {https://openreview.net/forum?id=hkujvAPVsg} {Hipporag: Neurobiologically inspired long-term memory for large language models}.
\newblock In \emph{The Thirty-eighth Annual Conference on Neural Information Processing Systems}.

\bibitem[{Gutiérrez et~al.(2025)Gutiérrez, Shu, Qi, Zhou, and Su}]{gutierrez2025hipporag2}
Bernal~Jiménez Gutiérrez, Yiheng Shu, Weijian Qi, Sizhe Zhou, and Yu~Su. 2025.
\newblock \href {https://arxiv.org/abs/2502.14802} {From rag to memory: Non-parametric continual learning for large language models}.
\newblock \emph{Preprint}, arXiv:2502.14802.

\bibitem[{Hou et~al.(2024)Hou, Tamoto, and Miyashita}]{10.1145/3613905.3650839}
Yuki Hou, Haruki Tamoto, and Homei Miyashita. 2024.
\newblock \href {https://doi.org/10.1145/3613905.3650839} {"my agent understands me better": Integrating dynamic human-like memory recall and consolidation in llm-based agents}.
\newblock In \emph{Extended Abstracts of the CHI Conference on Human Factors in Computing Systems}, CHI EA '24, New York, NY, USA. Association for Computing Machinery.

\bibitem[{Huet et~al.(2025)Huet, Houidi, and Rossi}]{huet2025episodic}
Alexis Huet, Zied~Ben Houidi, and Dario Rossi. 2025.
\newblock \href {https://openreview.net/forum?id=6ycX677p2l} {Episodic memories generation and evaluation benchmark for large language models}.
\newblock In \emph{The Thirteenth International Conference on Learning Representations}.

\bibitem[{Keil(1979)}]{keil1979semantic}
Frank~C Keil. 1979.
\newblock \emph{Semantic and conceptual development: An ontological perspective}.
\newblock Harvard University Press.

\bibitem[{Laird(2012)}]{10.5555/2222503}
John~E. Laird. 2012.
\newblock \emph{The Soar Cognitive Architecture}.
\newblock The MIT Press.

\bibitem[{Lee et~al.(2024)Lee, Chen, Furuta, Canny, and Fischer}]{pmlr-v235-lee24c-readagent}
Kuang-Huei Lee, Xinyun Chen, Hiroki Furuta, John Canny, and Ian Fischer. 2024.
\newblock \href {https://proceedings.mlr.press/v235/lee24c.html} {A human-inspired reading agent with gist memory of very long contexts}.
\newblock In \emph{Proceedings of the 41st International Conference on Machine Learning}, volume 235 of \emph{Proceedings of Machine Learning Research}, pages 26396--26415. PMLR.

\bibitem[{Li et~al.(2025)Li, Yang, Zhang, Deng, Wang, and Chua}]{li-etal-2025-hello}
Hao Li, Chenghao Yang, An~Zhang, Yang Deng, Xiang Wang, and Tat-Seng Chua. 2025.
\newblock \href {https://aclanthology.org/2025.naacl-long.272/} {Hello again! {LLM}-powered personalized agent for long-term dialogue}.
\newblock In \emph{Proceedings of the 2025 Conference of the Nations of the Americas Chapter of the Association for Computational Linguistics: Human Language Technologies (Volume 1: Long Papers)}, pages 5259--5276, Albuquerque, New Mexico. Association for Computational Linguistics.

\bibitem[{Maharana et~al.(2024)Maharana, Lee, Tulyakov, Bansal, Barbieri, and Fang}]{maharana-etal-2024-evaluating-locomo}
Adyasha Maharana, Dong-Ho Lee, Sergey Tulyakov, Mohit Bansal, Francesco Barbieri, and Yuwei Fang. 2024.
\newblock \href {https://doi.org/10.18653/v1/2024.acl-long.747} {Evaluating very long-term conversational memory of {LLM} agents}.
\newblock In \emph{Proceedings of the 62nd Annual Meeting of the Association for Computational Linguistics (Volume 1: Long Papers)}, pages 13851--13870, Bangkok, Thailand. Association for Computational Linguistics.

\bibitem[{Mandler(2014)}]{mandler2014stories}
Jean~Matter Mandler. 2014.
\newblock \emph{Stories, scripts, and scenes: Aspects of schema theory}.
\newblock Psychology Press.

\bibitem[{Mao et~al.(2022)Mao, Dou, and Qian}]{10.1145/3477495.3531961}
Kelong Mao, Zhicheng Dou, and Hongjin Qian. 2022.
\newblock \href {https://doi.org/10.1145/3477495.3531961} {Curriculum contrastive context denoising for few-shot conversational dense retrieval}.
\newblock In \emph{Proceedings of the 45th International ACM SIGIR Conference on Research and Development in Information Retrieval}, SIGIR '22, page 176–186, New York, NY, USA. Association for Computing Machinery.

\bibitem[{Martins et~al.(2022)Martins, Marinho, and Martins}]{martins-etal-2022-infinity-former}
Pedro~Henrique Martins, Zita Marinho, and Andre Martins. 2022.
\newblock \href {https://doi.org/10.18653/v1/2022.acl-long.375} {$\infty$-former: Infinite memory transformer}.
\newblock In \emph{Proceedings of the 60th Annual Meeting of the Association for Computational Linguistics (Volume 1: Long Papers)}, pages 5468--5485, Dublin, Ireland. Association for Computational Linguistics.

\bibitem[{Meylani(2024)}]{meylani2024innovations}
Rusen Meylani. 2024.
\newblock Innovations with schema theory: Modern implications for learning, memory, and academic achievement.
\newblock \emph{International Journal for Multidisciplinary Research}, 6(1):2582--2160.

\bibitem[{Murphy(2004)}]{murphy2004big}
Gregory Murphy. 2004.
\newblock \emph{The big book of concepts}.
\newblock MIT press.

\bibitem[{Ong et~al.(2025)Ong, Kim, Gwak, Chae, Kwon, Jo, Hwang, Lee, and Yeo}]{ong-etal-2025-towards}
Kai Tzu-iunn Ong, Namyoung Kim, Minju Gwak, Hyungjoo Chae, Taeyoon Kwon, Yohan Jo, Seung-won Hwang, Dongha Lee, and Jinyoung Yeo. 2025.
\newblock \href {https://aclanthology.org/2025.naacl-long.435/} {Towards lifelong dialogue agents via timeline-based memory management}.
\newblock In \emph{Proceedings of the 2025 Conference of the Nations of the Americas Chapter of the Association for Computational Linguistics: Human Language Technologies (Volume 1: Long Papers)}, pages 8631--8661, Albuquerque, New Mexico. Association for Computational Linguistics.

\bibitem[{{OpenAI}(2025)}]{openai2025gpt41}
{OpenAI}. 2025.
\newblock {GPT-4.1}.
\newblock \url{https://openai.com/index/gpt-4-1/}.
\newblock Accessed: 2025-05-17.

\bibitem[{Pan et~al.(2025)Pan, Wu, Jiang, Luo, Cheng, Li, Yang, Lin, Zhao, Qiu, and Gao}]{pan2025secom}
Zhuoshi Pan, Qianhui Wu, Huiqiang Jiang, Xufang Luo, Hao Cheng, Dongsheng Li, Yuqing Yang, Chin-Yew Lin, H.~Vicky Zhao, Lili Qiu, and Jianfeng Gao. 2025.
\newblock \href {https://openreview.net/forum?id=xKDZAW0He3} {Secom: On memory construction and retrieval for personalized conversational agents}.
\newblock In \emph{The Thirteenth International Conference on Learning Representations}.

\bibitem[{Piaget et~al.(1952)Piaget, Cook et~al.}]{piaget1952origins}
Jean Piaget, Margaret Cook, and 1 others. 1952.
\newblock \emph{The origins of intelligence in children}, volume~8.
\newblock International universities press New York.

\bibitem[{Qiu et~al.(2024)Qiu, Zhao, Ziser, Korhonen, Ponti, and Cohen}]{qiu-etal-2024-large}
Yifu Qiu, Zheng Zhao, Yftah Ziser, Anna Korhonen, Edoardo Ponti, and Shay Cohen. 2024.
\newblock \href {https://doi.org/10.18653/v1/2024.naacl-long.391} {Are large language model temporally grounded?}
\newblock In \emph{Proceedings of the 2024 Conference of the North American Chapter of the Association for Computational Linguistics: Human Language Technologies (Volume 1: Long Papers)}, pages 7064--7083, Mexico City, Mexico. Association for Computational Linguistics.

\bibitem[{Rumelhart(2017)}]{rumelhart2017schemata}
David~E Rumelhart. 2017.
\newblock Schemata: The building blocks of cognition.
\newblock In \emph{Theoretical issues in reading comprehension}, pages 33--58. Routledge.

\bibitem[{Rumelhart et~al.(1976)Rumelhart, Norman et~al.}]{rumelhart1976accretion}
David~E Rumelhart, Donald~A Norman, and 1 others. 1976.
\newblock \emph{Accretion, tuning and restructuring: Three modes of learning}.
\newblock Citeseer.

\bibitem[{Schacter and Addis(2007)}]{schacter2007constructive}
Daniel~L Schacter and Donna~Rose Addis. 2007.
\newblock On the constructive episodic simulation of past and future events.
\newblock \emph{Behavioral and Brain Sciences}, 30(3):331--332.

\bibitem[{Schacter and Tulving(1994)}]{schacter1994memory}
Daniel~L Schacter and Endel Tulving. 1994.
\newblock Memory systems 1994.
\newblock \emph{Memory Systems}, 199.

\bibitem[{Schank and Abelson(2013)}]{schank2013scripts}
Roger~C Schank and Robert~P Abelson. 2013.
\newblock \emph{Scripts, plans, goals, and understanding: An inquiry into human knowledge structures}.
\newblock Psychology press.

\bibitem[{Shan et~al.(2025)Shan, Luo, Zhu, Yuan, and Wu}]{shan2025cognitivememorylargelanguage}
Lianlei Shan, Shixian Luo, Zezhou Zhu, Yu~Yuan, and Yong Wu. 2025.
\newblock \href {https://arxiv.org/abs/2504.02441} {Cognitive memory in large language models}.
\newblock \emph{Preprint}, arXiv:2504.02441.

\bibitem[{Squire(1987)}]{squire1987memory}
L~Squire. 1987.
\newblock Memory and brain oxford university press: New york.

\bibitem[{Team et~al.(2025)Team, Kamath, Ferret, Pathak, Vieillard, Merhej, Perrin, Matejovicova, Ramé, Rivière, Rouillard, Mesnard, Cideron, bastien Grill, Ramos, Yvinec, Casbon, Pot, Penchev, Liu, Visin, Kenealy, Beyer, Zhai, Tsitsulin, Busa-Fekete, Feng, Sachdeva, Coleman, Gao, Mustafa, Barr, Parisotto, Tian, Eyal, Cherry, Peter, Sinopalnikov, Bhupatiraju, Agarwal, Kazemi, Malkin, Kumar, Vilar, Brusilovsky, Luo, Steiner, Friesen, Sharma, Sharma, Gilady, Goedeckemeyer, Saade, Feng, Kolesnikov, Bendebury, Abdagic, Vadi, György, Pinto, Das, Bapna, Miech, Yang, Paterson, Shenoy, Chakrabarti, Piot, Wu, Shahriari, Petrini, Chen, Lan, Choquette-Choo, Carey, Brick, Deutsch, Eisenbud, Cattle, Cheng, Paparas, Sreepathihalli, Reid, Tran, Zelle, Noland, Huizenga, Kharitonov, Liu, Amirkhanyan, Cameron, Hashemi, Klimczak-Plucińska, Singh, Mehta, Lehri, Hazimeh, Ballantyne, Szpektor, Nardini, Pouget-Abadie, Chan, Stanton, Wieting, Lai, Orbay, Fernandez, Newlan, yeong Ji, Singh, Black, Yu, Hui, Vodrahalli, Greff, Qiu,
  Valentine, Coelho, Ritter, Hoffman, Watson, Chaturvedi, Moynihan, Ma, Babar, Noy, Byrd, Roy, Momchev, Chauhan, Sachdeva, Bunyan, Botarda, Caron, Rubenstein, Culliton, Schmid, Sessa, Xu, Stanczyk, Tafti, Shivanna, Wu, Pan, Rokni, Willoughby, Vallu, Mullins, Jerome, Smoot, Girgin, Iqbal, Reddy, Sheth, Põder, Bhatnagar, Panyam, Eiger, Zhang, Liu, Yacovone, Liechty, Kalra, Evci, Misra, Roseberry, Feinberg, Kolesnikov, Han, Kwon, Chen, Chow, Zhu, Wei, Egyed, Cotruta, Giang, Kirk, Rao, Black, Babar, Lo, Moreira, Martins, Sanseviero, Gonzalez, Gleicher, Warkentin, Mirrokni, Senter, Collins, Barral, Ghahramani, Hadsell, Matias, Sculley, Petrov, Fiedel, Shazeer, Vinyals, Dean, Hassabis, Kavukcuoglu, Farabet, Buchatskaya, Alayrac, Anil, Dmitry, Lepikhin, Borgeaud, Bachem, Joulin, Andreev, Hardin, Dadashi, and Hussenot}]{gemmateam2025gemma3technicalreport}
Gemma Team, Aishwarya Kamath, Johan Ferret, Shreya Pathak, Nino Vieillard, Ramona Merhej, Sarah Perrin, Tatiana Matejovicova, Alexandre Ramé, Morgane Rivière, Louis Rouillard, Thomas Mesnard, Geoffrey Cideron, Jean bastien Grill, Sabela Ramos, Edouard Yvinec, Michelle Casbon, Etienne Pot, Ivo Penchev, and 197 others. 2025.
\newblock \href {https://arxiv.org/abs/2503.19786} {Gemma 3 technical report}.
\newblock \emph{Preprint}, arXiv:2503.19786.

\bibitem[{Thakur et~al.(2021)Thakur, Reimers, R{\"u}ckl{\'e}, Srivastava, and Gurevych}]{thakur2021beir}
Nandan Thakur, Nils Reimers, Andreas R{\"u}ckl{\'e}, Abhishek Srivastava, and Iryna Gurevych. 2021.
\newblock \href {https://openreview.net/forum?id=wCu6T5xFjeJ} {{BEIR}: A heterogeneous benchmark for zero-shot evaluation of information retrieval models}.
\newblock In \emph{Thirty-fifth Conference on Neural Information Processing Systems Datasets and Benchmarks Track (Round 2)}.

\bibitem[{Wang et~al.(2024{\natexlab{a}})Wang, Zhang, Yang, Chen, Tang, Zhang, Chen, Lin, Song, Zhao, Xu, Dou, Wang, and Wen}]{wang2024userbehaviorsimulationlarge}
Lei Wang, Jingsen Zhang, Hao Yang, Zhiyuan Chen, Jiakai Tang, Zeyu Zhang, Xu~Chen, Yankai Lin, Ruihua Song, Wayne~Xin Zhao, Jun Xu, Zhicheng Dou, Jun Wang, and Ji-Rong Wen. 2024{\natexlab{a}}.
\newblock \href {https://arxiv.org/abs/2306.02552} {User behavior simulation with large language model based agents}.
\newblock \emph{Preprint}, arXiv:2306.02552.

\bibitem[{Wang et~al.(2025)Wang, Fu, Cao, Wang, Tian, and Ding}]{WANG2025130193}
Qingyue Wang, Yanhe Fu, Yanan Cao, Shuai Wang, Zhiliang Tian, and Liang Ding. 2025.
\newblock \href {https://doi.org/10.1016/j.neucom.2025.130193} {Recursively summarizing enables long-term dialogue memory in large language models}.
\newblock \emph{Neurocomputing}, 639:130193.

\bibitem[{Wang et~al.(2024{\natexlab{b}})Wang, Han, Wu, He, Zhou, Sadeq, Chen, He, Wang, Haffari et~al.}]{wang2024towards}
Yu~Wang, Chi Han, Tongtong Wu, Xiaoxin He, Wangchunshu Zhou, Nafis Sadeq, Xiusi Chen, Zexue He, Wei Wang, Gholamreza Haffari, and 1 others. 2024{\natexlab{b}}.
\newblock Towards lifespan cognitive systems.
\newblock \emph{arXiv preprint arXiv:2409.13265}.

\bibitem[{Wu et~al.(2025{\natexlab{a}})Wu, Wang, Yu, Zhang, Chang, and Yu}]{wu2025longmemeval}
Di~Wu, Hongwei Wang, Wenhao Yu, Yuwei Zhang, Kai-Wei Chang, and Dong Yu. 2025{\natexlab{a}}.
\newblock \href {https://openreview.net/forum?id=pZiyCaVuti} {Longmemeval: Benchmarking chat assistants on long-term interactive memory}.
\newblock In \emph{The Thirteenth International Conference on Learning Representations}.

\bibitem[{Wu et~al.(2025{\natexlab{b}})Wu, Liang, Zhang, Wang, Zhang, Guo, Tang, and Liu}]{wu2025humanmemoryaimemory}
Yaxiong Wu, Sheng Liang, Chen Zhang, Yichao Wang, Yongyue Zhang, Huifeng Guo, Ruiming Tang, and Yong Liu. 2025{\natexlab{b}}.
\newblock \href {https://arxiv.org/abs/2504.15965} {From human memory to ai memory: A survey on memory mechanisms in the era of llms}.
\newblock \emph{Preprint}, arXiv:2504.15965.

\bibitem[{Xu et~al.(2025)Xu, Liang, Mei, Gao, Tan, and Zhang}]{xu2025a-mem}
Wujiang Xu, Zujie Liang, Kai Mei, Hang Gao, Juntao Tan, and Yongfeng Zhang. 2025.
\newblock A-mem: Agentic memory for llm agents.
\newblock \emph{arXiv preprint arXiv:2502.12110}.

\bibitem[{Yang et~al.(2024)Yang, Yang, Zhang, Hui, Zheng, Yu, Li, Liu, Huang, Wei, Lin, Yang, Tu, Zhang, Yang, Yang, Zhou, Lin, Dang, Lu, Bao, Yang, Yu, Li, Xue, Zhang, Zhu, Men, Lin, Li, Xia, Ren, Ren, Fan, Su, Zhang, Wan, Liu, Cui, Zhang, and Qiu}]{qwen2.5}
An~Yang, Baosong Yang, Beichen Zhang, Binyuan Hui, Bo~Zheng, Bowen Yu, Chengyuan Li, Dayiheng Liu, Fei Huang, Haoran Wei, Huan Lin, Jian Yang, Jianhong Tu, Jianwei Zhang, Jianxin Yang, Jiaxi Yang, Jingren Zhou, Junyang Lin, Kai Dang, and 22 others. 2024.
\newblock Qwen2.5 technical report.
\newblock \emph{arXiv preprint arXiv:2412.15115}.

\bibitem[{Yuan et~al.(2025)Yuan, Sun, Li, Wang, Cao, and Li}]{yuan-etal-2025-personalized}
Ruifeng Yuan, Shichao Sun, Yongqi Li, Zili Wang, Ziqiang Cao, and Wenjie Li. 2025.
\newblock \href {https://aclanthology.org/2025.coling-main.254/} {Personalized large language model assistant with evolving conditional memory}.
\newblock In \emph{Proceedings of the 31st International Conference on Computational Linguistics}, pages 3764--3777, Abu Dhabi, UAE. Association for Computational Linguistics.

\bibitem[{Zhang et~al.(2025)Zhang, Li, Zeng, and Wang}]{zhang2025jasperstelladistillationsota}
Dun Zhang, Jiacheng Li, Ziyang Zeng, and Fulong Wang. 2025.
\newblock \href {https://arxiv.org/abs/2412.19048} {Jasper and stella: distillation of sota embedding models}.
\newblock \emph{Preprint}, arXiv:2412.19048.

\bibitem[{Zhong et~al.(2024)Zhong, Guo, Gao, Ye, and Wang}]{Zhong_Guo_Gao_Ye_Wang_2024}
Wanjun Zhong, Lianghong Guo, Qiqi Gao, He~Ye, and Yanlin Wang. 2024.
\newblock \href {https://doi.org/10.1609/aaai.v38i17.29946} {Memorybank: Enhancing large language models with long-term memory}.
\newblock \emph{Proceedings of the AAAI Conference on Artificial Intelligence}, 38(17):19724--19731.

\bibitem[{Zhu et~al.(2025)Zhu, Xie, Liu, Li, and Hu}]{zhu2025knowledgegraphguidedretrievalaugmented}
Xiangrong Zhu, Yuexiang Xie, Yi~Liu, Yaliang Li, and Wei Hu. 2025.
\newblock \href {https://arxiv.org/abs/2502.06864} {Knowledge graph-guided retrieval augmented generation}.
\newblock \emph{Preprint}, arXiv:2502.06864.

\end{thebibliography}
